\documentclass{article}
\usepackage[T1]{fontenc}
\usepackage{iclr2027_conference,times}
\usepackage{amsmath,amssymb}
\usepackage{booktabs,tabularx,array}
\usepackage{graphicx,algorithm,algpseudocode,wrapfig}
\usepackage{needspace}
\usepackage{multirow}
\usepackage{enumitem}
\algrenewcommand{\algorithmicrequire}{\textbf{Input:}}
\algrenewcommand{\algorithmicensure}{\textbf{Output:}}  
\usepackage{xcolor}
\usepackage{hyperref}
\usepackage{url}
\definecolor{topcolor}{HTML}{E7DCEF}
\definecolor{secondcolor}{HTML}{E5EDF3}
\hypersetup{hidelinks,  colorlinks=true,
  citecolor=blue,
  linkcolor=black,
  urlcolor=black,pdfauthor={Linji Hao, Yanji He, Zihan Wang, Hansi Yang, Jiayuan Zhang, Hongyan Hao, Qi Gu, Yan Feng, Han Wu},pdftitle={Interlock: From Local Correctness to Task Completion in Long-Horizon Agents}}

\DeclareRobustCommand{\Interlock}{\mbox{\textsc{Interlock}}}

\usepackage{textcomp,listings,tcolorbox}
\tcbuselibrary{skins,breakable,listings}
\newcommand{\InterlockField}[1]{\texttt{\detokenize{#1}}}
\newtcblisting{InterlockPromptBox}[1]{
  enhanced,breakable,listing only,listing engine=listings,
  width=\linewidth,colback=white,colframe=black,
  colbacktitle=black,coltitle=white,
  boxrule=0.5pt,arc=1.5mm,boxsep=0.8mm,
  left=1.8mm,right=1.8mm,top=1.5mm,bottom=1.5mm,
  toptitle=1mm,bottomtitle=1mm,titlerule=0pt,
  fonttitle=\normalfont\bfseries,
  title={#1},title after break={#1\ (continued)},
  before skip=7pt,after skip=9pt,
  listing options={
    language={},basicstyle=\ttfamily\footnotesize,
    columns=fullflexible,keepspaces=true,
    showstringspaces=false,showspaces=false,showtabs=false,
    breaklines=true,breakatwhitespace=false,breakindent=0pt,
    numbers=none,aboveskip=0pt,belowskip=0pt,tabsize=2,upquote=true
  }
}
\title{\Interlock{}: From Local Correctness to Task Completion in Long-Horizon Agents}
\author{%
  \parbox[t]{\dimexpr\textwidth-2\tabcolsep\relax}{%
    \centering
    \mbox{Linji Hao$^{1}$\thanks{Equal contribution.}}\quad
    \mbox{Yanji He$^{2}$\footnotemark[1]}\quad
    \mbox{Zihan Wang$^{3}$\footnotemark[1]}\quad
    \mbox{Hansi Yang$^{4}$}\quad
    \mbox{Jiayuan Zhang$^{4}$}\\
    \mbox{Hongyan Hao$^{4}$}\quad
    \mbox{Qi Gu$^{4}$\thanks{Corresponding authors.}}\quad
    \mbox{Yan Feng$^{4}$}\quad
    \mbox{Han Wu$^{1}$\footnotemark[2]}\\[0.5em]
    \normalfont
    $^{1}$Peking University \quad
    $^{2}$The Hong Kong University of Science and Technology (Guangzhou)\\
    $^{3}$Tsinghua University\quad
    $^{4}$Meituan, China\\
    \texttt{haolinji26@stu.pku.edu.cn} \quad \texttt{guqi03@meituan.com} \quad 
  }%
}

\AddToHookNext{shipout/foreground}{%
    \begin{tikzpicture}[remember picture, overlay]
        \node[
            anchor=north,
            inner sep=0pt,
            yshift=-0.40in
        ] at (current page.north) {%
            \begin{minipage}{\dimexpr\paperwidth-1.5in\relax}
                \raggedright
                \includegraphics[width=3.6cm]
                    {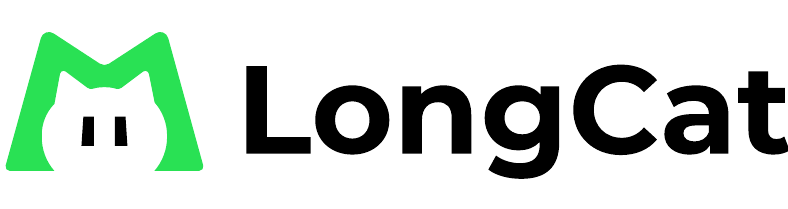}\par
                \vspace{3pt}
                \hrule height 0.4pt
            \end{minipage}%
        };
    \end{tikzpicture}%
}

\iclrfinalcopy
\begin{document}
\maketitle
\lhead{}
\renewcommand{\headrulewidth}{0pt}
\begin{abstract}
Large language model agents increasingly undertake long-horizon tasks in complex environments characterized by fragmented evidence, partially completed states, and dependencies across resources and artifacts. Existing agents may satisfy individual requirements while failing to maintain these relationships, producing locally correct but incomplete outcomes. These failures arise because task completion depends on relationships among environmental evidence, consequential decisions, and required effects. To address this completion gap, we propose \textbf{\Interlock{}}, a framework for task synthesis that represents each task as a typed completion graph. Grounding edges connect public evidence to decisions, coupling hyperedges impose joint constraints across components, and realization edges connect decisions to all required downstream effects. \Interlock{} compiles this graph into a resettable workspace, a public request, and an evaluation rubric. A source-coverage review checks whether the original requirements survive generation, while a public-input review checks whether the rubric is supported by the materials available to the solver. An agent-as-judge inspects execution artifacts and runs verification scripts, retaining trajectories judged to satisfy all mandatory requirements. To evaluate training utility, we fine-tune two Qwen3.6 backbones on their own accepted trajectories. Across the two backbones, training improves AutomationBench task pass rates by 4.00–4.50 percentage points, JobBench weighted scores by 3.20–6.64 points, and Harvey LAB criterion pass rates by 3.89–4.16 percentage points. These results demonstrate the training utility of the synthesized experience across external task environments.
\end{abstract}

\section{Introduction}\label{introduction}

Long-horizon agents increasingly work with records, rules, resources, and partially completed artifacts. A short request can require reconciling conflicting evidence, determining which objects need attention, selecting a feasible plan, and updating several related results. The environment supplies the facts and constraints that determine this work. A decision can also activate further obligations: deferring an action may require an explanation and a handoff, while changing an allocation may require corresponding revisions to schedules and budgets. These obligations can outlive the interaction in which they are discovered: after many intervening reads, computations, and edits, the agent must still apply the earlier evidence and decisions to every affected result.

\begin{figure}[t]
\centering
\includegraphics[width=\linewidth]{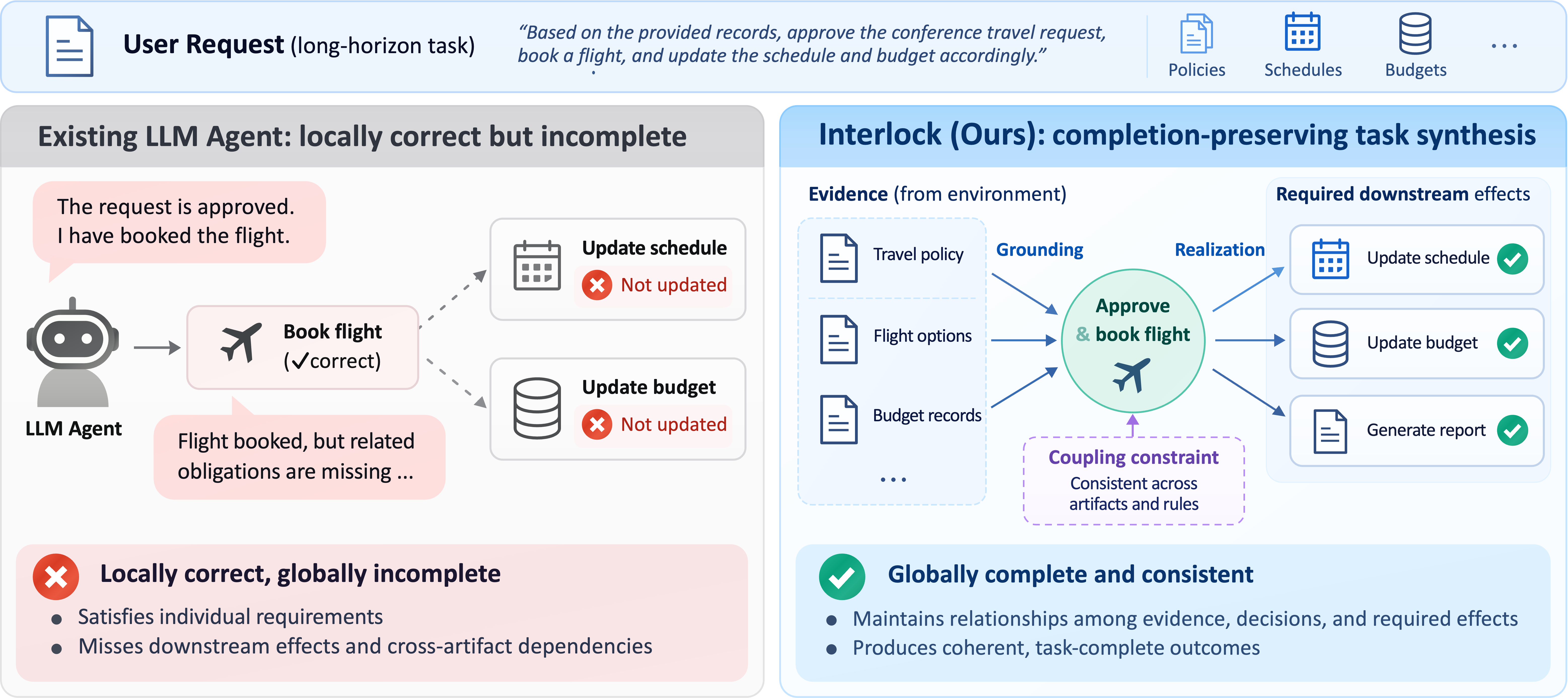}
\caption{\textbf{From local correctness to task completion.}
A correct flight booking can still leave schedule and budget updates
unfinished. \Interlock{} preserves grounding, coupling, and realization
relations during task synthesis to connect evidence, decisions, and
required downstream effects.}
\label{fig:intro-motivation}
\end{figure}

Local progress can coexist with an incomplete outcome (Figure~\ref{fig:intro-motivation}). A correct decision can leave necessary consequences unfulfilled, and mutually consistent outputs can share an unsupported assumption. These failures concern the relationships that connect evidence, decisions, and the requested outcome. They motivate our central question: \emph{how can task synthesis preserve conditional obligations and turn their fulfillment into transferable training experience?}

We characterize this structure as a typed completion graph. \emph{Grounding} connects environmental evidence and rules to the decisions they support. \emph{Coupling} links components that must share a feasible choice or satisfy a joint constraint. \emph{Realization} connects a decision to every downstream record, artifact, or state that must reflect it. Their interaction is especially demanding over long executions: an early finding can constrain work completed many interactions later, and a revised decision can require coordinated changes to several existing artifacts.

Task synthesis and environment generation provide a scalable basis for agent training. Prior work constructs verified tool-use data~\citep{qin2023toolllm,liu2024apigen}, generates environments and task scenarios~\citep{hu2025agentgen,song2026envscaler}, and composes extended tasks through tool dependencies and reusable workflows~\citep{tu2026scaleenv,shi2025taskcraft,li2026rst}. Common specifications can also align instructions, environments, and verifiers~\citep{ivanov2026anchor}. Our organizing unit is the complete set of conditional obligations for an outcome. It determines both the unresolved work placed in an environment and the evidence needed to accept the result.

We propose \textbf{\Interlock{}} to make this graph an explicit synthesis constraint. Reusable domain descriptions and archetype contracts define evidence roles, task-specific decisions and shared states, joint conditions, and downstream consumers. Three generation roles compile the same graph into a workspace, task/contract, and rubric while leaving concrete associations and decisions to the agent. A source-coverage review traces every typed relation from the original graph through these artifacts; a public-input critic independently checks whether the resulting criteria are supported by solver-visible materials and assessable from execution artifacts. Completion-filtered training then selects executions that realize every applicable effect.

Our evaluation connects task structure, execution behavior, and training utility. Statistical analyses and case studies examine dependencies and completion failures. The training protocol uses Qwen3.6-27B and Qwen3.6-35B-A3B separately, retaining each model's own accepted rollouts for SFT of that same model. Both backbones improve on AutomationBench, JobBench Main and Easy, and Harvey LAB. Gains over the original checkpoints reach 4.00--4.50 percentage points in AutomationBench task pass rate, 3.20--6.64 points in JobBench weighted score, and 3.89--4.16 percentage points in Harvey LAB criterion pass rate. These results show that experience from complete synthetic executions transfers to independently authored tasks across different environments. Our contributions are threefold:
\begin{itemize}[leftmargin=*]
    \item \textbf{A typed representation of task completion.} We model long-horizon tasks as completion graphs whose grounding, coupling, and realization relations connect public evidence, consequential decisions, joint constraints, and required effects.
    \item \textbf{Graph-grounded synthesis and review.} We compile the same graph into the environment, task specification, deliverable contract, and outcome criteria, with complementary reviews for source coverage and public support.
    \item \textbf{Empirical validation of training utility.} We evaluate the generated experience through completion-filtered fine-tuning of two backbones and measure transfer across three external benchmarks.
\end{itemize}
\section{Related Work}\label{sec:related_work}
Task synthesis and environment generation provide a scalable source of agent experience.
ToolAlpaca~\citep{tang2023toolalpaca}, ToolLLM~\citep{qin2023toolllm}, and ToolACE~\citep{liu2025toolace} broaden tool coverage and interaction diversity, while APIGen~\citep{liu2024apigen} incorporates format, execution, and semantic verification.
Beyond tool-use data, AgentGen~\citep{hu2025agentgen}, AutoForge~\citep{cai2025autoforge}, and EnvScaler~\citep{song2026envscaler} extend synthesis to environments and their associated planning tasks or executable scenarios.
Agent-World~\citep{dong2026agent} further couples environment-task discovery with self-evolving training, and ClawGym~\citep{bai2026clawgym} brings synthesis to persistent workspaces through persona-driven intents, skill-grounded operations, and hybrid verification.
To construct extended interactions, ToolFlow~\citep{wang2025toolflow} plans coherent tool-use dialogues, while APIGen-MT~\citep{prabhakar2025apigenmt} turns verified action blueprints into multi-turn interactions.
BUTTON~\citep{chen2025button}, TaskCraft~\citep{shi2025taskcraft}, and AgentSynth~\citep{xie2025agentsynth} construct complex requests by composing or expanding smaller tasks.
Complementary approaches, including VOYAGER~\citep{wang2023voyager}, OS-Genesis~\citep{sun2025osgenesis}, and Explorer~\citep{pahuja2025explorer}, obtain curricula or tasks through environmental state and exploration.
Alongside task complexity, another line of work addresses consistency among generated instructions, environments, and verifiers.
Endless Terminals~\citep{gandhi2026endless} generates containerized tasks with completion tests, while RST~\citep{li2026rst} extends verified reference solutions and realigns instructions and verifiers with the resulting workflows.
CodeMidas~\citep{ye2026codemidas} derives coding tasks from existing implementations, grounds synthesized tests in execution of the original code, and filters tasks through execution checks and agent rollouts.
Anchor~\citep{ivanov2026anchor} addresses artifact drift by deriving task artifacts and certified solutions from a common constraint program.
Graph-guided synthesis provides more explicit control over the dependencies underlying generated tasks.
ScaleEnv~\citep{tu2026scaleenv} constructs tasks from tested tool dependencies, EnvFactory~\citep{xu2026envfactory} uses topology-aware sampling to generate grounded queries, and SkillSynth~\citep{fan2026skillsynth} instantiates paths through a scenario-mediated skill graph.
These approaches organize generation around tool connectivity and skill relationships.
\Interlock{} instead uses task-completion relationships as synthesis constraints: its typed completion graph grounds decisions in evidence, imposes joint constraints, and specifies required downstream effects.

\section{Methodology}
\label{method}

\begin{figure}[t]
\centering
\includegraphics[width=\linewidth]{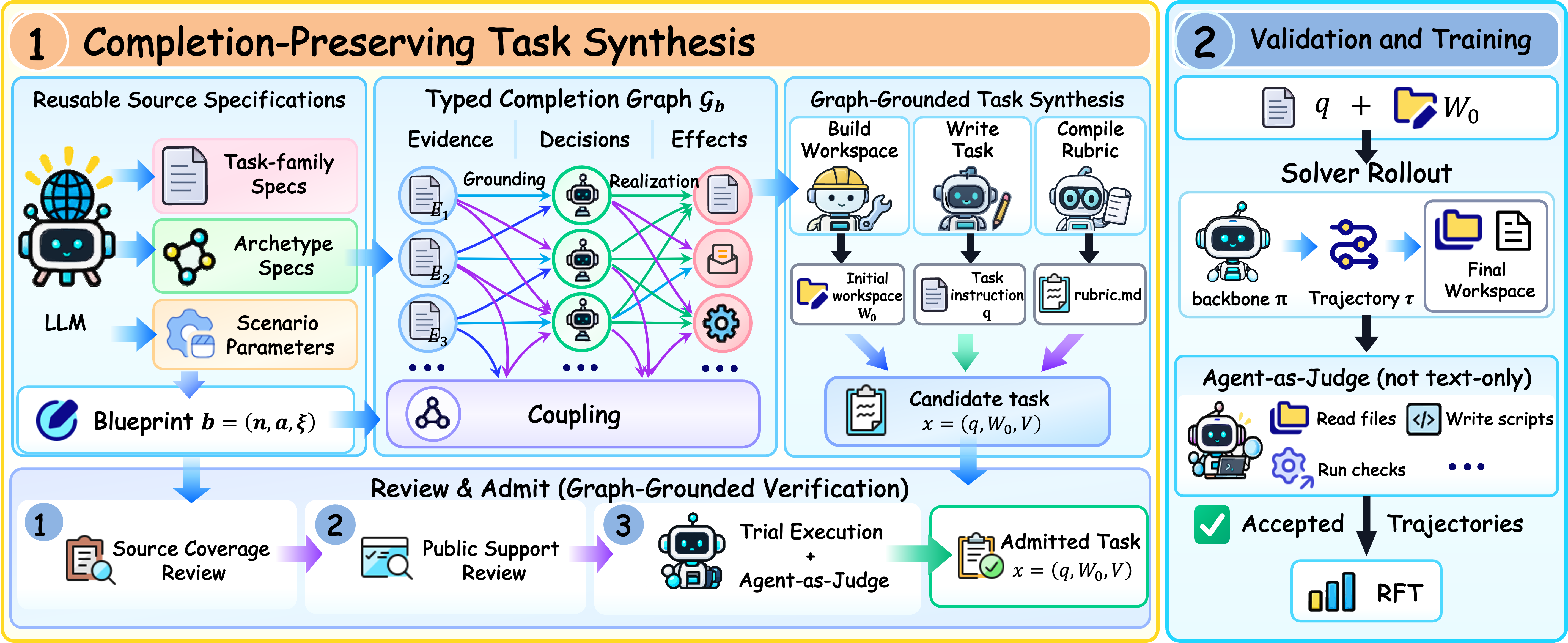}
\caption{\textbf{\Interlock{} overview.}
\textbf{Left:} Grounding connects evidence to decision variables;
Coupling imposes joint constraints across decisions and effects;
Realization specifies the required consequences of each decision.
These relations guide construction of the workspace $W_0$, public
request $q$, and private rubric $V$, followed by source-coverage
and public-support review.
\textbf{Right:} Real execution and an agent-as-judge provide evidence
for task admission and trajectory selection. Each backbone is
fine-tuned on its own accepted rollouts. 
}
\label{fig:method-overview}
\end{figure}

\Interlock{} uses a special data structure, \emph{typed completion graph}, to preserve
the relationships among environmental evidence, agent decisions,
and required effects throughout task construction and evaluation.
Figure~\ref{fig:method-overview} outlines the workflow.
Its left panel follows the graph through workspace construction,
task writing, rubric compilation, and complementary source-coverage
and public-support reviews. Its right panel shows real execution,
outcome evaluation, and the collection of accepted trajectories
for training. Specialized agents receive the inputs designated
for their roles; Algorithm~\ref{alg:interlock} in Appendix~\ref{app:algorithm} specifies
task admission and the subsequent training procedure.

A task is a package $x=(q,W_0,V)$, where $q$ is the user
request , $W_0$ is the initial workspace, 
and $V$ is the structured private rubric.
The agent receives only $(q,W_0)$, 
while the evaluation rubric $V$ remains outside the agent's context
and is used for requirement review and outcome evaluation.

\subsection{Typed Completion Graph}
\label{sec:completion-graph}

\paragraph{Nodes and relations.}

Specifically, the typed completion graph records the requirements for an outcome:
\begin{equation}
\mathcal G_b =
\bigl(
\mathcal E_b,\mathcal D_b,\mathcal Y_b,
\mathcal R_b^G,\mathcal R_b^C,\mathcal R_b^R
\bigr).
\label{eq:completion-graph}
\end{equation}
Here, $b$ identifies the task specification used to construct the graph.
$\mathcal E_b$ contains evidence and rule roles, such as an authoritative
record or an applicable policy. $\mathcal D_b$ contains decisions and
shared states that the solver must determine. $\mathcal Y_b$ contains
required \emph{effects}: observable results such as a revised record,
a completed artifact, or an updated workspace state. The three relation
sets are
\begin{equation}
\begin{aligned}
\mathcal R_b^G
&\subseteq \mathcal E_b \times \mathcal D_b,
\mathcal R_b^C
&\subseteq 2^{\mathcal D_b\cup\mathcal Y_b}\times\Psi_b,
\mathcal R_b^R
&\subseteq \mathcal D_b \times \mathcal Y_b,
\end{aligned}
\label{eq:relation-types}
\end{equation}
where $\Psi_b$ contains the public conditions that govern joint
feasibility or consistency.

\textbf{Grounding} specifies which evidence and rules govern a decision.
It covers the decision's factual basis and any applicable
source-precedence or authorization rule.
\textbf{Coupling} specifies a condition that several decisions or
effects must satisfy together. A coupling relation $(S,\psi)$ connects
a set of components $S$ through a shared condition $\psi$. It is a
\emph{hyperedge} because the condition can involve more than two
components.
\textbf{Realization} specifies which effects must reflect a decision.
Together, the three types determine whether the decisions are supported,
their uses are jointly valid, and their required consequences are
delivered.

\paragraph{Conditional effects.}

Each effect $y$ carries four fields,
$y=(B_y,\Omega_y,\alpha_y,\phi_y)$.
$B_y$ identifies the public evidence or rule that establishes the
obligation. $\Omega_y$ specifies the source population over which the
obligation must be considered. $\alpha_y$ states when it applies to an
object in that population, and $\phi_y$ states what must hold in the
delivered outcome. An object may be an employee, an invoice, a resource,
or another task-specific entity.





\subsection{Relation-Constrained Task Synthesis}
\label{sec:practice}

\paragraph{Reusable specifications.}

A \emph{task domain} groups tasks sharing a business or professional
context. A \emph{blueprint} $b=(n,a,\xi)$ combines a task family $n$
within a domain, an archetype $a$, and scenario parameters $\xi$.
A task family supplies a reusable
objective and its obligation set $\mathcal O_n$. An archetype supplies
a recurring relationship pattern $\rho_a$, such as several outputs
sharing a decision or several allocations sharing a capacity
constraint. Scenario parameters instantiate the entities, source
facts, and initial records. The graph is constructed as
\begin{equation}
\mathcal G_b =
\operatorname{BuildCompletionGraph}
\bigl(\mathcal O_n,\rho_a,\xi\bigr).
\label{eq:build-completion-graph}
\end{equation}
Evidence requirements produce evidence nodes, decisions produce decision
nodes, and obligations produce conditional effects. Shared conditions
produce coupling relations. Different archetypes induce different
relation patterns. The graph records the requirements declared for the
selected task family.

\paragraph{Task artifacts.}

Three specialized agents translate the graph
into an executable task package.
The operators below denote their respective generation stages,
with the resulting artifacts passed to subsequent agents:

\begin{equation}
\begin{aligned}
(W_0,I,\mu_W)
&= \operatorname{BuildWorkspace}(b,\mathcal G_b),\\
(q,H,\mu_T)
&= \operatorname{WriteTask}(b,\mathcal G_b,W_0,I),\\
(V,\mu_V)
&= \operatorname{CompileRubric}(\mathcal G_b,q,W_0,I,H).
\end{aligned}
\label{eq:task-construction}
\end{equation}
The environment builder creates $W_0$ and an \emph{evidence index} $I$
that locates the relevant public files and records. The task setter
creates the request $q$ and a private \emph{deliverable contract} $H$:
a structured inventory of required outputs and their conditions.
Every obligation in $H$ must be supported by $q$ or the public
workspace. The rubric compiler creates criteria for the decisions,
joint conditions, and required effects.

The \emph{carrier maps} $\mu_W$, $\mu_T$, and $\mu_V$ record where a
graph requirement appears in each artifact. A carrier is a concrete
location, such as a policy clause, a request statement, a contract
entry, or an evaluation criterion. These maps provide references for
review, rather than solved assignments. We retain the construction
record
$c=(b,\mathcal G_b,x,I,H,\mu)$,
with $\mu=(\mu_W,\mu_T,\mu_V)$.

\paragraph{Construction constraints.}

Each relation type places a concrete requirement on generation.
For grounding, the controlling sources must be present and the decision
must remain unresolved. For coupling, the shared condition must
constrain the combined decisions or outputs. For realization, every
declared downstream recipient or artifact must have an observable
required effect and an outcome check.

Source conflicts, partial records, and stale outputs instantiate work
that the solver must resolve through these relationships. The rubric
retains their public conditions, including the rules that determine
which objects require an effect.

\subsection{Requirement Review and Task Admission}
\label{completion-preserving-rubric-compilation}

\paragraph{Source coverage.}

The source-coverage review starts from the original obligation
specification and archetype requirements. It checks their representation
in $\mathcal G_b$ and then traces the graph through the workspace,
task/contract, and rubric.

Let
$\Gamma_b=\mathcal R_b^G\cup\mathcal R_b^C\cup\mathcal R_b^R$,
and let
$\mathcal S(\gamma)\subseteq
\{W,T,V\}$
specify the stages in which relation $\gamma$ must appear. Define
\begin{equation}
P(c)=P_{\mathrm{spec}}(c)
\prod_{\gamma\in\Gamma_b}
\prod_{s\in\mathcal S(\gamma)}P_s(\gamma).
\label{eq:relation-preservation}
\end{equation}
$P_{\mathrm{spec}}(c)=1$ when the graph covers the original
specification. $P_s(\gamma)=1$ when the stage contains the required
carrier and preserves the relation's public basis, applicability,
scope, and outcome condition. The review passes when $P(c)=1$.

Structural checks establish reference integrity and semantic review
assesses the actual requirements and their representations. Carrier
maps locate the material to inspect. This source anchor keeps an
obligation in view even when both the generated request and rubric
omit it.

\paragraph{Public support.}
The public-input critic receives only $(q,W_0,V)$. It checks that
the criteria follow from the public request and workspace, that their
joint requirements are consistent, and that outcomes can be judged
from the specified evidence.
Source coverage and public support therefore check complementary
directions: the first preserves the original requirements, while
the second establishes a public basis for the delivered evaluation.
Both reviews return findings with evidence references and generation
roles perform any required revisions.

\paragraph{Execution-based admission.}

After review, a validation policy executes the candidate from a fresh
copy of $W_0$. The judge examines the initial workspace, the final
workspace, and their changes. It returns criterion judgments and a
separate package-quality assessment.
Admission requires both reviews to pass, a normally terminated
validation execution, complete judgments, and a sound package
assessment. It does not require that the validation solver complete
the task: task admission retains valid challenges, while trajectory
selection identifies accepted executions.
Confirmed defects return to their producing stages, and revised
packages receive fresh reviews. Rubric-only repairs may reuse a
normally ended execution when the public inputs are unchanged while
other repairs require a new execution. Revisions are bounded and
versioned.
Algorithm~\ref{alg:interlock} groups these review, repair, and
validation steps as $\operatorname{ReviewAndAdmit}$, which returns
an admitted task or $\varnothing$.

\subsection{Completion-Filtered Self-Supervision}
\label{learning-from-scored-rollouts}

\paragraph{Rubric-based evaluation.}

Each generated task includes a private \texttt{rubric\_spec.json} file
specifying its outcome criteria and verification metadata, together with
its readable rendering \texttt{rubric.md}.
We use an \emph{agent-as-judge} with terminal access
to evaluate each rollout $\tau$ on task $x$.
Rather than scoring a fixed text response, the judge
actively gathers and checks evidence: guided by the rubric
and execution results, it inspects relevant files and,
when needed, writes and executes verification scripts.
These checks assess the final artifacts and workspace changes
against the public request, initial workspace, and rubric.
The judge returns a pass/fail judgment for each required criterion.
We set $A_V(x,\tau)=1$ when all required criteria pass
and any separately declared quality threshold is met,
otherwise, $A_V(x,\tau)=0$.
Throughout the paper, an \emph{accepted trajectory} is one with
$A_V(x,\tau)=1$. Accepted trajectories are retained for training.

\paragraph{Per-backbone training.}

We use \emph{rejection-sampling fine-tuning} (RFT):
each backbone generates candidate executions, and supervised
fine-tuning uses its accepted subset.

For backbone $k$, the initial policy $\pi_0^{(k)}$ makes $M$
attempts per admitted task, each from a reset workspace.
Let $\mathcal T_{\mathrm{ok}}^{(k)}$ contain attempts with
normal termination, complete judgments, and no confirmed
package defect. Its training set is
$\mathcal D_{\mathrm{RFT}}^{(k)}
=
\bigl\{
(x,\tau)\in\mathcal T_{\mathrm{ok}}^{(k)}:
A_V(x,\tau)=1
\bigr\}.
\label{eq:training-set}$
Starting from $\pi_0^{(k)}$, we optimize
\begin{equation}
\mathcal L_{\mathrm{RFT}}^{(k)}(\theta_k)
=
-\mathbb E_{(x,\tau)\sim\mathcal D_{\mathrm{RFT}}^{(k)}}
\left[
\sum_{t\in\mathcal J(\tau)}
\log\pi_{\theta_k}(z_t\mid q,h_t)
\right].
\label{eq:rft-loss}
\end{equation}
$\theta_k$ denotes the trainable policy parameters,
$\mathcal J(\tau)$ the supervised assistant-token positions,
$z_t$ the target token, and $h_t$ the preceding interaction
history. Supervision covers assistant reasoning, responses, and tool calls;
workspace observations remain conditioning context.



\suppressfloats[t]
\section{Experiments}\label{experiments}

\begin{table}[t]
\caption{Main results on unseen benchmark tasks. AutomationBench and Harvey LAB report task and criterion pass rates (\%), respectively; JobBench Main and Easy report weighted scores (0--100). Each panel compares training conditions within one backbone. Higher is better, and the best score within each backbone is shown in bold.}
\label{tab:results}
\vspace{2px}
\centering\small
\setlength{\tabcolsep}{4pt}
\begin{tabularx}{\linewidth}{@{}Xc|c|cc|c@{}}
\toprule
\multirow{2}{*}{Base Model} & \multirow{2}{*}{Method} & \multirow{2}{*}{\textbf{AutomationBench}} & \multicolumn{2}{c|}{\textbf{JobBench}} & \multirow{2}{*}{\textbf{Harvey LAB}} \\
& & & Main & Easy & \\
\midrule
& Original & 14.33 & 14.11 & 54.46 & 73.96 \\
{\textbf{Qwen3.6-27B}} & ClawGym-SFT & 9.00 & 11.82 & 48.01 & 75.03 \\
& \textbf{Interlock-RFT} & \textbf{18.33} & \textbf{18.46} & \textbf{61.10} & \textbf{77.85} \\
\midrule
& Original & 9.00 & 11.81 & 51.86 & 54.18 \\
{\textbf{Qwen3.6-35B-A3B}}  & ClawGym-SFT & 5.17 & 6.25 & 40.66 & 54.97 \\
& \textbf{Interlock-RFT} & \textbf{13.50} & \textbf{15.01} & \textbf{56.17} & \textbf{58.34} \\
\bottomrule
\end{tabularx}
\end{table}

\begin{figure}[t]
\centering
\includegraphics[width=0.245\linewidth]{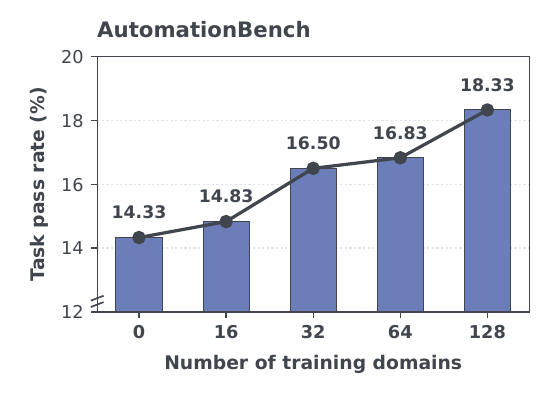}\hfill%
\includegraphics[width=0.245\linewidth]{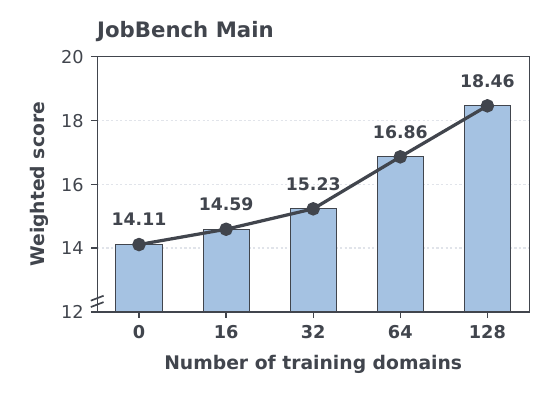}\hfill%
\includegraphics[width=0.245\linewidth]{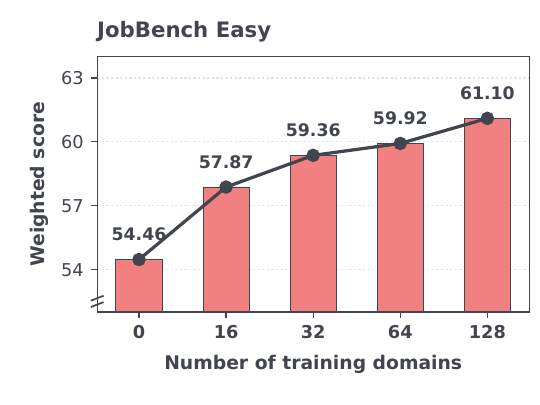}\hfill%
\includegraphics[width=0.245\linewidth]{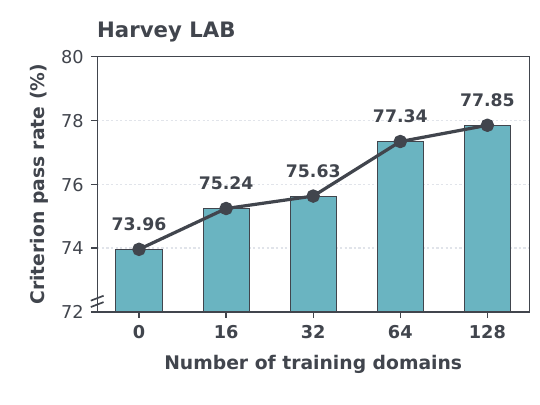}
\caption{\textbf{Broader domain coverage improves transfer at a fixed trajectory count.} The original checkpoint
is labeled 0, and 128 denotes all domains in this model's training set
for the main experiment.
AutomationBench reports task pass rate (\%),
JobBench Main and Easy report weighted scores (0--100), and
Harvey LAB reports criterion pass rate (\%). Lines connect the
evaluated settings.}
\label{fig:task-diversity}
\end{figure}

\subsection{Experimental Setup}\label{sec:experimental-setup}

\paragraph{Models and training.}
We use \textbf{Qwen3.6-27B} and \textbf{Qwen3.6-35B-A3B} as the base models and consider the following three settings: the original checkpoint, ClawGym-SFT on released
demonstrations~\citep{bai2026clawgym}, and Interlock-RFT on each
backbone's own accepted rollouts.
Both backbones use thinking mode during training and evaluation.
The reported corpus retains 3,278 tasks with at least one accepted
rollout from each backbone. Five attempts per task and backbone yield 32,780 attempts,
of which 32,370 terminate normally.
Each Interlock-RFT model uses 8,957 of its own accepted trajectories and
approximately $154\mathrm{M}$ supervised tokens, including reasoning,
responses, and tool calls.

\paragraph{Benchmarks and metrics.}
We evaluate on AutomationBench (600 public tasks)~\citep{shepard2026automationbench},
JobBench Main and Easy (65 tasks each)~\citep{li2026jobbench}, and
Harvey LAB (1,251 public tasks)~\citep{harveylab2026}.
These evaluation instances are excluded from synthesis and RFT.
The respective metrics are task pass rate, task-averaged weighted score,
and criterion pass rate, all on a 0--100 scale with higher values better.
JobBench runs in OpenCode, and GPT-5.5 judges both JobBench and Harvey LAB.
Appendix~\ref{app:experimental-setup} gives the detailed training and
evaluation protocols and metric definitions; Appendix~\ref{app:task-statistics}
reports statistics of the retained corpus.

\subsection{Main Results}\label{main-results}

Interlock-RFT achieves the highest score in every evaluation column for both backbones (Table~\ref{tab:results}), outperforming the original checkpoints and ClawGym-SFT. Each backbone is trained on its own accepted trajectories, and the gains extend across coordinated API execution, professional deliverables, and legal work.
For Qwen3.6-27B, \Interlock{} improves AutomationBench task pass rate by 4.00 percentage points, JobBench Main and Easy weighted scores by 4.35 and 6.64 points, and Harvey LAB criterion pass rate by 3.89 percentage points over the original checkpoint. For Qwen3.6-35B-A3B, the gains are 4.50 percentage points on AutomationBench, 3.20 and 4.31 score points on JobBench Main and Easy, and 4.16 percentage points on Harvey LAB. The improvements therefore cover task completion on AutomationBench, weighted rubric fulfillment on JobBench, and criterion satisfaction on Harvey LAB.



\subsection{Task Diversity and Transfer}\label{sec:task-diversity}

We vary the number of training domains over 16, 32, 64, and 128.
All four Interlock-RFT settings use 8,957 training trajectories,
matching the Qwen3.6-27B main experiment.
Tasks are grouped by domain, with each domain containing task families
that share a business or professional context (Section~\ref{sec:practice}).
Domain coverage therefore measures the breadth of professional settings
represented in training.
Figure~\ref{fig:task-diversity} compares Qwen3.6-27B under these
Interlock-RFT settings with its original checkpoint, labeled 0.
The setting covering all 128 domains corresponds to the Interlock-RFT result in
Table~\ref{tab:results}.
Performance increases at every successive setting on all four metrics,
with 128 domains achieving the highest values.
In \Interlock{}, expanding domain coverage varies the professional evidence,
objectives, and constraints over which grounding, coupling, and realization
must be satisfied. At a fixed trajectory count, broader domain coverage
supplies more varied instances of these completion requirements.
The shared gains across API execution, professional deliverables,
and legal work support learning transferable completion behavior
from complete executions of diverse tasks.

\subsection{Agentic Verification for Completion-Filtered Training}
\label{sec:judge-ablation}

We compare two trajectory judges on Qwen3.6-27B to isolate the
training contribution of the trajectory judge.
Both judging configurations use the same judge backbone, public task
requests, rubrics, and acceptance rule
(Section~\ref{learning-from-scored-rollouts}) to filter the fixed rollout
pool underlying the Qwen3.6-27B Interlock-RFT result.
The text-only LLM-as-judge receives a fixed textual representation of
the initial and final workspaces and their diff, while the agent-as-judge
can inspect the same snapshots and write and execute verification scripts.
The agent-filtered condition reuses the main Interlock-RFT checkpoint.
The text-filtered condition is fine-tuned from the same initial
Qwen3.6-27B checkpoint with the same  optimization settings.
Verification output counts all judge-generated tokens, including
verification-script and tool-call tokens, summed over all verification
calls for each candidate rollout and averaged over the entire pool.
Tool-returned content is excluded from this output count.

\begin{table}[t]
\centering
\small
\setlength{\tabcolsep}{4pt}
\renewcommand{\arraystretch}{1.12}
\caption{\textbf{Agentic verification analysis.}
Both filters use the same judge backbone, rubrics, and Qwen3.6-27B
rollout pool. The resulting models share initialization and SFT settings.
Judge output sums all verification calls and is averaged over all
candidate rollouts, including accepted and rejected ones; tool returns
are excluded ($\mathrm{k}=10^3$ tokens).}
\label{tab:judge-ablation}
\begin{tabular}{@{}lccccc@{}}
\toprule
\textbf{Trajectory filter}
& \shortstack{\textbf{Judge output} $\downarrow$\\\textbf{(k tokens/rollout)}}
& \shortstack{\textbf{Automation}\\\textbf{Bench} $\uparrow$}
& \shortstack{\textbf{JobBench}\\\textbf{Main} $\uparrow$}
& \shortstack{\textbf{JobBench}\\\textbf{Easy} $\uparrow$}
& \shortstack{\textbf{Harvey LAB}\\$\uparrow$} \\
\midrule
Original (no SFT) & -- & 14.33 & 14.11 & 54.46 & 73.96 \\
LLM-as-judge & 11.03
& 13.17 & 15.14 & 56.53 & 75.28 \\
Agent-as-judge & 12.28
& \textbf{18.33} & \textbf{18.46} & \textbf{61.10} & \textbf{77.85} \\
\bottomrule
\end{tabular}
\end{table}

Agent-based filtering improves all four downstream metrics at the
matched training budget (Table~\ref{tab:judge-ablation}).
Average judge output increases from approximately 11.03k to 12.28k
tokens per candidate rollout, an 11.3\% increase, or about 1.25k
additional tokens.
These gains support checking submitted artifacts through inspection
and recomputation when selecting demonstrations for task completion.
With the rollout pool, judge backbone, and supervision budget held
fixed, executable verification improves the training value of the
selected trajectories. 


\subsection{Long-Horizon Execution on Synthesized Tasks}\label{sec:execution-horizons}

The 3,278 retained tasks each receive five rollout attempts from
Qwen3.6-27B and five from Qwen3.6-35B-A3B, giving 32,780 attempts.
Excluding 410 non-normal executions leaves 32,370 trajectories for
execution-horizon analysis. Each backbone selects 8,957 of its own
accepted trajectories for training,
with approximately $154\mathrm{M}$ supervised tokens in each training set.
We characterize execution horizon over this normally terminated pool through
assistant turns, tool calls, and recorded trajectory tokens.

\begin{figure}[t]
\centering
\includegraphics[width=\linewidth]{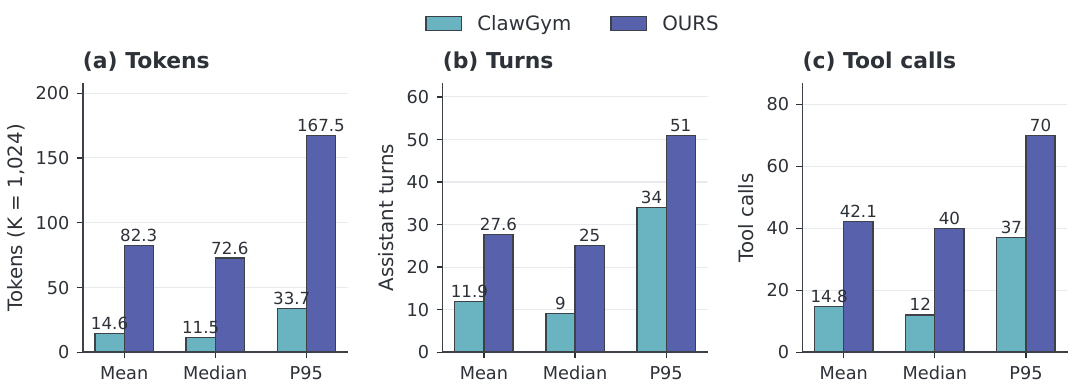}
\caption{\textbf{Execution horizons of \Interlock{} and ClawGym trajectories.}
Bars report the mean, median, and 95th percentile (P95) of
recorded tokens, assistant turns, and tool calls. Each trajectory has equal weight, and
$\mathrm{K}=1{,}024$ tokens.
Appendix~\ref{app:trajectory-statistics} gives the measurement protocol.}
\label{fig:trajectory-horizon-summary}
\end{figure}

Figure~\ref{fig:trajectory-horizon-summary} compares the 32,370
normally terminated \Interlock{} trajectories with 24,497 released ClawGym trajectories.
\Interlock{} trajectories average 84,293 recorded tokens, 27.63 assistant
turns, and 42.10 tool calls, compared with 14,940 tokens, 11.92 turns,
and 14.76 calls for ClawGym.
The corresponding ratios are $5.64\times$, $2.32\times$, and
$2.85\times$. The difference extends to the centers of the
distributions: median assistant turns are 25 versus 9, and median
tool calls are 40 versus 12. Appendix~\ref{app:trajectory-statistics}
specifies the population and counting rules and provides the cumulative
distributions and histograms.
Assistant turns measure successive model responses, while tool calls
measure the actions submitted during those responses. Their joint
increase grounds the long-horizon characterization in repeated
interaction with the task environment. These execution statistics
show that \Interlock{} produces tasks that elicit sustained agent work,
providing experience in carrying evidence, decisions, and required
effects across many interactions.

\Needspace{26\baselineskip}
\subsection{Generalization across Task Environments}\label{sec:generalization-realism}

\begin{wrapfigure}{r}{0.43\textwidth}
\centering
\includegraphics[width=\linewidth]{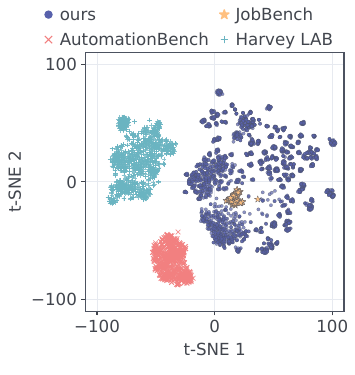}
\small
\caption{\textbf{Task-statement embeddings.}
Joint PCA--t-SNE projection of normalized BGE-M3 embeddings;
``ours'' denotes \Interlock{}.}
\label{fig:task-embeddings}
\end{wrapfigure}

Figure~\ref{fig:task-embeddings} visualizes selected public task
statements from \Interlock{}, AutomationBench, JobBench Main, and
Harvey LAB. We encode each statement with BGE-M3~\citep{chen2024m3}
and project the normalized embeddings into two dimensions using
PCA followed by t-SNE~\citep{vandermaaten2008tsne}.

\Interlock{} tasks span a broad area of the projection and form multiple
dispersed groups, reflecting diverse task content. This breadth extends
to environment construction: the retained corpus covers 128 domains
and 484 task families, with multi-file workspaces containing task-specific
records, policies, and unfinished artifacts
(Appendix~\ref{app:task-statistics}). These settings vary the evidence,
joint constraints, and downstream obligations over which grounding,
coupling, and realization must be satisfied.

AutomationBench, JobBench, and Harvey LAB together evaluate
out-of-distribution (OOD) transfer to independently authored tasks
with their own materials, interfaces, and outcome requirements.
Their task statements form dataset-specific groups in the projection,
and their evaluation instances are excluded from synthesis and RFT.
Consistent gains across all three benchmarks and both backbones
(Table~\ref{tab:results}) demonstrate the training value of complete
synthetic executions beyond the generated task distribution.
Together with the domain-diversity results
(Figure~\ref{fig:task-diversity}), they support learning transferable
completion behavior by connecting evidence, decisions, and required
effects across varied professional settings.



\subsection{Case Studies of Task Completion}\label{sec:synthesized-case-studies}

\begin{figure}[t]
\centering
\includegraphics[width=0.49\linewidth]{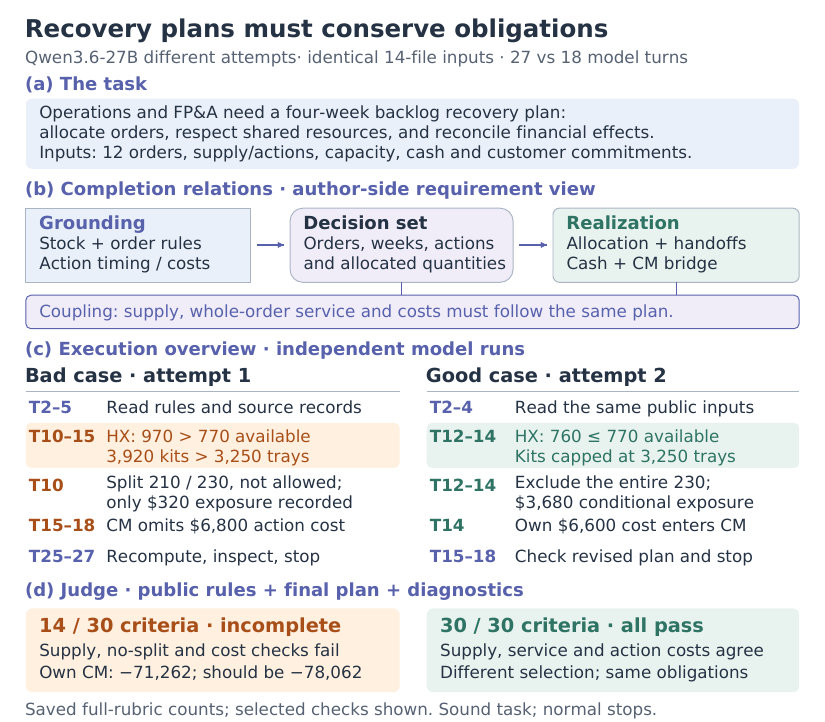}\hfill%
\includegraphics[width=0.49\linewidth]{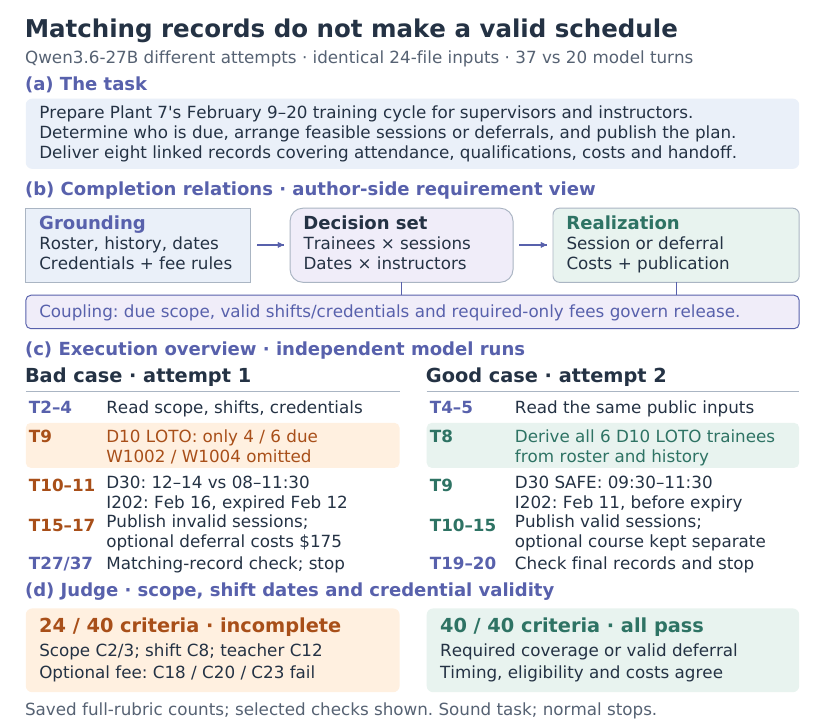}
\caption{\textbf{Case study.}
Left: recovery planning couples supply, whole-order service, and action
costs. Right: training schedules couple due scope, shift windows, and
dated qualifications.}
\label{fig:synthesized-cases-main}
\end{figure}

Figure~\ref{fig:synthesized-cases-main} traces completion requirements
in two synthesized tasks using public inputs and recorded artifacts.
For each task, we compare two rollout attempts of Qwen3.6-27B on
identical public inputs.

\paragraph{Operating recovery.}
A four-week recovery plan must reconcile supply, order commitments,
and financial effects. In attempt 1, Qwen3.6-27B allocates 970 HX units against
770 available; advancing an inbound shipment changes its arrival time,
not the total supply. Its 3,920-kit plan also exceeds 3,250 shared trays,
partially serves an unsplittable 230-unit order, and omits a selected
\$6,800 expedite cost from contribution margin despite recording the
cash outflow. In attempt 2, the same model uses 760 HX units, caps kits at
3,250, excludes the whole unsplittable order, and deducts its own
selected action costs.

\paragraph{Training schedules.}
The February 2026 training cycle links due employees, shift availability,
instructor credentials, and publication records. Attempt 1 of Qwen3.6-27B produces
matching schedule and publication rows but omits two of six due
trainees, schedules a 12:00--14:00 class outside an 08:00--11:30 release
window, and assigns an instructor on February 16 whose credential
expires on February 12. Attempt 2 of the same model retains all six
trainees, uses a shared 09:30--11:30 window, and schedules that
instructor on February 11. Appendix~\ref{app:synthesized-case-studies} extends the analysis to
field-trial protocols, charging networks, and payroll close.

\section{Conclusion}
\Interlock{} organizes long-horizon task synthesis around typed completion graphs. Grounding, coupling, and realization relations constrain how public evidence, consequential decisions, joint conditions, and required effects are compiled into task environments and outcome criteria. Dual-view review checks both graph preservation and public support, while completion-filtered RFT selects executions that realize every applicable graph effect. Training improves both Qwen3.6-27B and Qwen3.6-35B-A3B across AutomationBench, JobBench Main and Easy, and Harvey LAB, supporting the value of complete synthetic execution experience for transfer across task environments.
\clearpage
\label{end-main-text}
\section*{AI use statement}
We used generative AI tools to assist with prompt design and manuscript
preparation, including language polishing, bilingual editing, and the
preparation of figures and captions. As described in
Section~\ref{method}, AI agents are integral to the task synthesis and
training-data collection procedure: they generate initial workspaces,
task instructions, and outcome rubrics; review task requirements and
quality; execute tasks; and assess the resulting artifacts. The roles
and verification procedures for these agents are documented in the
methodology and experimental setup. We have manually reviewed all
AI-assisted work, including manuscript revisions, prompts, generated
task materials, and execution and evaluation records. We take
responsibility for the final content of this work, including its text,
analyses, claims, and artifacts produced with AI assistance.
\bibliography{references}
\bibliographystyle{iclr2027_conference}
\clearpage
\appendix
\section{Acceptance and Outcome Scoring}\label{app:outcome-scoring}

\paragraph{Obligation representation.}
Let $\mathcal R_x$ be the full set of mandatory obligations for task $x$. An obligation $r\in\mathcal R_x$ consists of its public evidence $B_r$, binary applicability condition $\alpha_r$, binary outcome condition $\phi_r$, and object scope $\Omega_r$:
\begin{equation}
r=(B_r,\alpha_r,\phi_r,\Omega_r),\qquad
\alpha_r(q,W_0,W_1^\tau)=1
\ \Longrightarrow\
\phi_r(W_0,W_1^\tau)=1.
\label{eq:conditional-requirement}
\end{equation}
Here, $\alpha_r$ determines whether the obligation applies given the request, initial materials, and delivered choices; $\phi_r$ determines whether its required effect holds; and $\Omega_r$ specifies the objects over which applicability and fulfillment are evaluated. We write $\alpha_r(\tau)$ and $\phi_r(\tau)$ as shorthand for the full arguments. Unconditional obligations have $\alpha_r=1$. The typed completion graph shares these obligations across its evidence, decision, and effect nodes without instantiating their correct values.




\paragraph{Training-set construction.}
For a sound task $x$, let $\mathcal T_x^{(k)}$ contain attempts generated by starting policy $\pi_0^{(k)}$, and let $\operatorname{normal}(\tau)$ equal one when execution ends normally and zero otherwise. The proposed retained set is
\begin{equation}
\mathcal D_{\mathrm{RFT}}^{(k)}
=\{(x,\tau):\tau\in\mathcal T_x^{(k)},\
\operatorname{normal}(\tau)=1,\
A_V(x,\tau)=1\}.
\label{eq:selected-distribution}
\end{equation}
Each backbone retains native messages and tool schemas from its own executions. Private rubrics, correspondence records, and judge feedback remain outside the training conversation.

\subsection{Diagnostic Outcome Scoring}\label{app:diagnostic-scoring}

The following equations define its partial-credit outcome statistics. Trajectory acceptance follows $A_V(x,\tau)=1$ as defined in Section~\ref{learning-from-scored-rollouts}.

For criterion $i$, the judge reports initial and final satisfaction $b_i,m_i\in\{0,1\}$. Declared support dependencies $D_i$ yield
\begin{equation}
\widetilde m_i=m_i\prod_{j\in D_i}\widetilde m_j.
\label{eq:effective-criterion}
\end{equation}
The extracted criterion-dependency graphs contain no cycles or unknown references. An acyclic dependency graph is the precondition for this recursion. These \texttt{depends\_on} edges propagate criterion credit; they are distinct from the grounding, coupling, and realization relations that define task semantics. Applicability and exceptions are expressed in each public criterion proposition.

Let $N_g$ contain the scored criteria in group $g$. A change criterion satisfied both initially and finally is excluded from $N_g$; preservation criteria remain scored. Let $E_g\subseteq N_g$ contain the group's essential criteria and let $\mathcal G_+$ contain groups with scored criteria. Then
\begin{equation}
\begin{aligned}
p_g&=\frac{1}{|N_g|}\sum_{i\in N_g}\widetilde m_i,
&s_g&=p_g\prod_{i\in E_g}\widetilde m_i,\\
S_V&=\frac{\sum_{g\in\mathcal G_+}w_gs_g}
{\sum_{g\in\mathcal G_+}w_g}.
\end{aligned}
\label{eq:outcome}
\end{equation}
Critical groups account for at most half the rubric weight. Essential failures zero their groups. The score $S_V$ summarizes partial fulfillment, while $A_V$ requires every mandatory obligation to be fulfilled and any separately declared quality threshold to be met.

\section{Behavioral Diagnostic Protocol}\label{app:behavioral-diagnostics}

A decision unit contains a publicly justified decision and the full bundle of applicable effects for its consumers. Let $u_i$ indicate a correct upstream decision and $e_i$ indicate fulfillment of every applicable downstream effect. The eligible units have at least one applicable downstream obligation. Follow-through is
\begin{equation}
\mathrm{FTR}=\frac{\sum_i u_i e_i}{\sum_i u_i}.
\label{eq:follow-through}
\end{equation}
The report includes $\sum_i u_i$, upstream correctness, and unconditional joint success over all eligible units. Outcomes are established from raw criteria and delivered artifacts. A paired analysis additionally reports units whose upstream decisions are correct in both checkpoints; resampling uses task families when tasks share a family and tasks otherwise, accounting for multiple units within an execution. An empty conditional denominator is reported as not applicable.

The diagnostic set is frozen outside the training data. A composition-transfer extension reserves selected domain--relationship combinations while retaining their constituent domains and relationships in other combinations. Relevant environment interventions change a public fact that affects applicability or feasibility; semantic-preserving controls reorder equivalent material. Each task version has its own quality review and permits all outcomes satisfying its public requirements. Measurements cover joint success across paired versions, correct updates to affected effects, and preservation of unaffected conditions.
\section{Trajectory Statistics and Comparison}\label{app:trajectory-statistics}

\paragraph{Rollout population and training selection.}
The experimental corpus retains 3,278 tasks on which both
Qwen3.6-27B and Qwen3.6-35B-A3B have at least one accepted
rollout. Five attempts per task and backbone give 32,780 attempts.
Excluding 410 abnormally terminated or incomplete executions leaves
32,370 normally terminated trajectories, which define the \Interlock{}
population in the execution-horizon figures.
The ClawGym comparison uses all 24,497 supplied release trajectories.
Appendix~\ref{app:experimental-setup} gives the selection protocol.
Each backbone independently selects 8,957 accepted trajectories
from its own normally terminated rollouts for its own RFT training,
using the acceptance rule in Section~\ref{learning-from-scored-rollouts}.
Each training set contains approximately $154\mathrm{M}$ supervised
tokens; the two model-specific sets contain 17,914 trajectories in total.

\paragraph{Measurement.}
An assistant turn is one recorded assistant message, including the
final response. Several tool calls in one assistant message count
as one turn and as multiple calls. HTTP retries and error responses
do not add assistant turns. We use the same Qwen3.6-27B tokenizer
for both corpora, with truncation and padding disabled. Token counts
sum recorded user, assistant, and tool content, call arguments,
saved reasoning, and one copy of the tool schema. Duplicate copies
of saved reasoning are reconciled before counting. These counts
measure recorded content, with each event counted once; they do
not sum repeatedly submitted conversation contexts. Inline ClawGym tool-call text is counted once as assistant
content. All distribution statistics weight trajectories equally
within their respective corpora.

The supervised-token budget counts assistant positions that contribute
to the training loss, including reasoning, responses, and tool calls.
Input-only messages, tool observations, and tool schemas are excluded.
The approximately $154\mathrm{M}$ tokens per backbone therefore measure
supervision in its 8,957 training trajectories; the execution-length
figures measure recorded interaction content over the 32,370 normally
terminated rollouts.
Here, $\mathrm{M}=10^6$, while the figure axes use
$\mathrm{K}=1{,}024$ recorded tokens.

\begin{figure}[t]
\centering
\includegraphics[width=\linewidth]{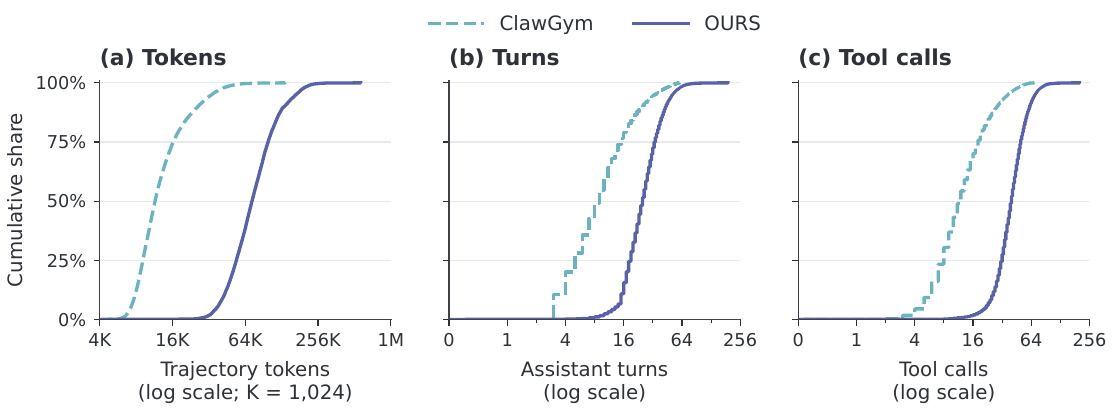}
\caption{\textbf{Cumulative execution-horizon distributions.}
Horizontal axes show recorded trajectory tokens, assistant turns,
and tool calls. The vertical axis is the fraction of trajectories
at or below each horizontal value; a curve farther to the right
indicates larger values at the same percentile.
The curves use the same 32,370 \Interlock{} and 24,497 ClawGym
trajectories and token-counting rules as
Figure~\ref{fig:trajectory-horizon-summary}; $\mathrm{K}=1{,}024$.}
\label{fig:trajectory-horizon-ecdf}
\end{figure}

\begin{figure}[t]
\centering
\includegraphics[width=\linewidth]{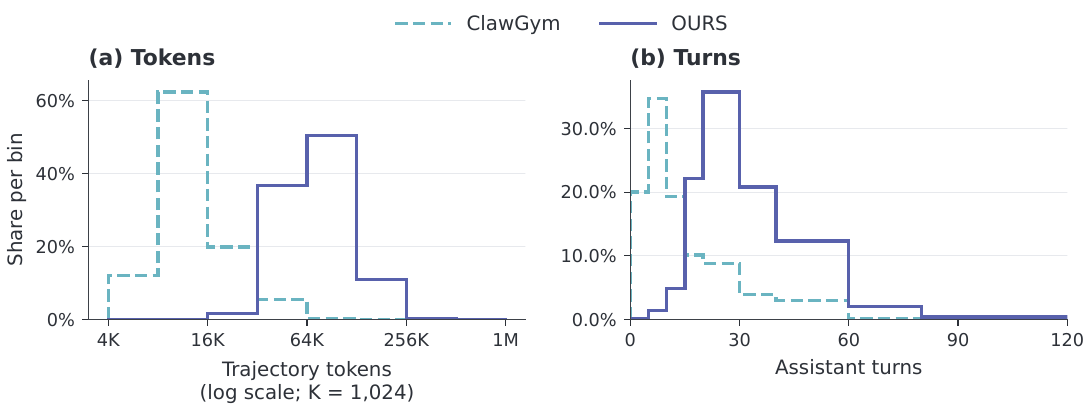}
\caption{\textbf{Trajectory-length and assistant-turn histograms.}
Each vertical value is the fraction of trajectories in the
corresponding bin. The two corpora share bin boundaries within
each panel; the plotted values are bin shares rather than
probability densities. The panels retain the full observed range
of the 32,370 \Interlock{} and 24,497 ClawGym trajectories, with the
same counting rules as Figure~\ref{fig:trajectory-horizon-summary};
$\mathrm{K}=1{,}024$ tokens.}
\label{fig:trajectory-horizon-histograms}
\end{figure}

\paragraph{Execution-horizon statistics.}
The \Interlock{} mean is 84,292.51 recorded tokens, 27.63 assistant turns,
and 42.10 tool calls per trajectory. The corresponding medians are
74,384.5 tokens, 25 turns, and 40 calls; the 95th percentiles are
171,549.1 tokens, 51 turns, and 70 calls.
ClawGym means are 14,939.62 tokens, 11.92 turns, and 14.76 calls,
yielding mean ratios of $5.64\times$, $2.32\times$, and $2.85\times$.
These values use the same population as
Figure~\ref{fig:trajectory-horizon-summary}.
Figures~\ref{fig:trajectory-horizon-ecdf} and
\ref{fig:trajectory-horizon-histograms} show the cumulative
distributions and bin shares for these two rollout pools.

\clearpage
\section{Statistics of the Retained Task Corpus}\label{app:task-statistics}

We analyze the 3,278 synthesized tasks retained after requiring at least
one accepted rollout from each backbone
(Appendix~\ref{app:experimental-setup}). The retained corpus covers
128 domains and 484 task families, spanning eight task archetypes.
Task families are defined within domains; archetypes describe recurring
task structures across domains (Section~\ref{sec:practice}).
Each instance contributes once under its recorded dominant
archetype. File counts consistently refer to the initial source workspace,
including materials, drafts, and templates. Rubric statistics count each
criterion once, without applying scoring weights; verification methods
refer to the methods declared in those criteria.

\begin{figure}[htbp]
\centering
\includegraphics[width=\linewidth]{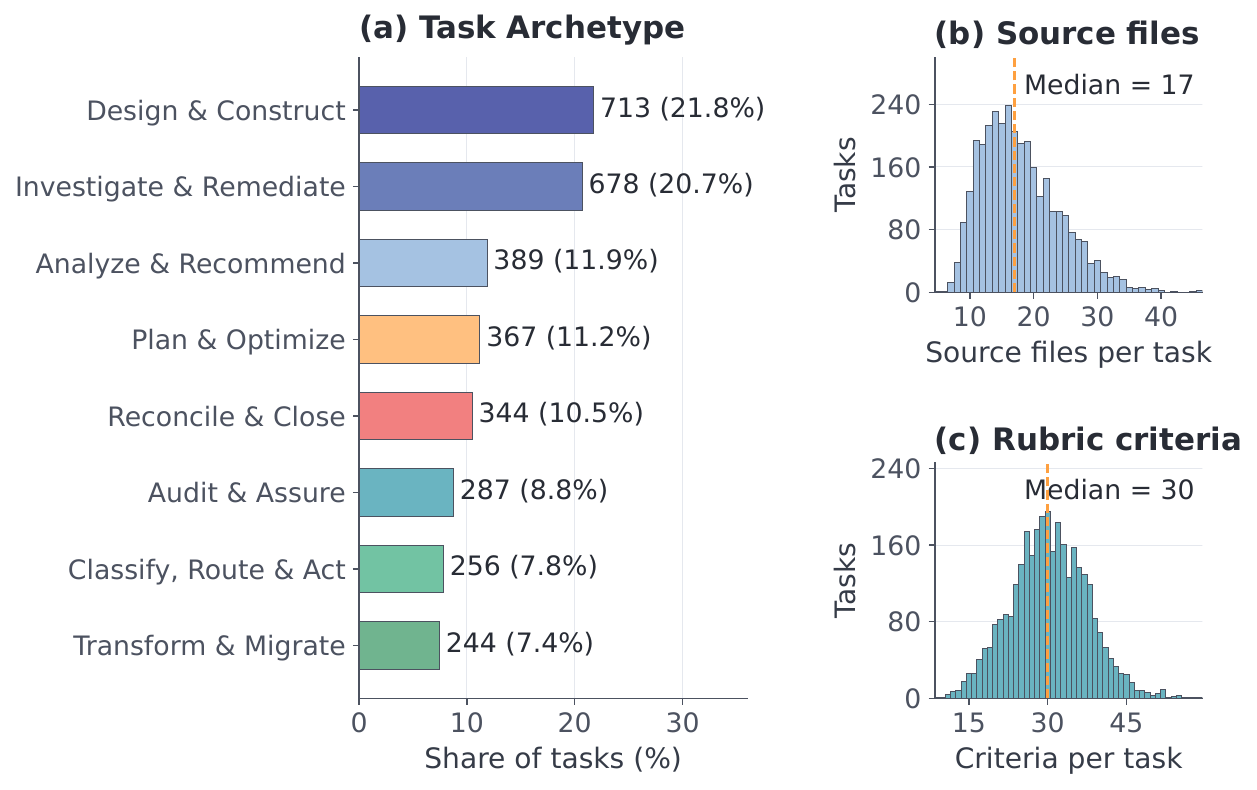}
\caption{\textbf{Eight task archetypes with multi-file workspaces and multi-criterion rubrics.}
All panels cover the 3,278 retained tasks: (a) archetype shares, (b) source workspace file counts,
and (c) rubric criterion counts. Orange dashed lines mark medians.}
\label{fig:task-archetype-overview}
\end{figure}

Figure~\ref{fig:task-archetype-overview} summarizes task composition and
the two artifacts that connect evidence to evaluation: the initial
workspace and the outcome rubric. The median task contains 17 source
files and 30 criteria. In \Interlock{}, the workspace supplies the public
evidence and rules that ground decisions, while the rubric specifies
conditions on those decisions, their joint consistency, and their
required effects. The following figures describe how these materials
and checks are distributed across the full corpus.

\subsection{Verification Methods}

\begin{figure}[htbp]
\centering
\includegraphics[width=\linewidth]{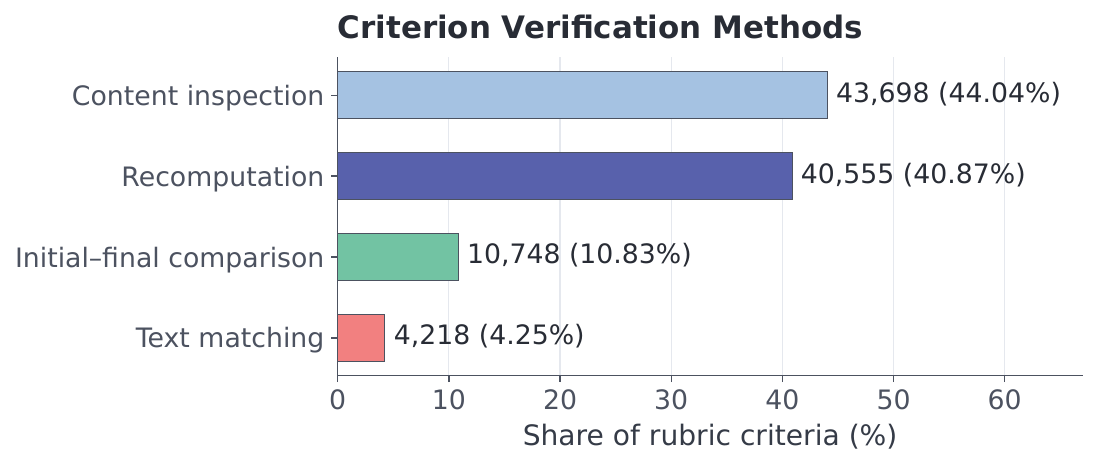}
\caption{\textbf{Content inspection and recomputation dominate the declared checks.}
Bars show counts and shares of all 99,219 criteria, with each criterion
counted once under its declared verification method.}
\label{fig:verification-method-bar}
\end{figure}

\paragraph{Criterion composition.}
Figure~\ref{fig:verification-method-bar} shows that content inspection
and recomputation account for 44.04\% and 40.87\% of criteria.
Initial--final comparison contributes 10.83\%, and text matching
contributes 4.25\%. These methods provide complementary ways to
evaluate completion: inspecting delivered content, checking
derived results, comparing changes with the initial state, and
verifying specified text. Their combination provides evidence
for evaluating supported decisions and required outcomes within
the same rubric.

\subsection{Verification Across Archetypes}

\begin{figure}[htbp]
\centering
\includegraphics[width=\linewidth]{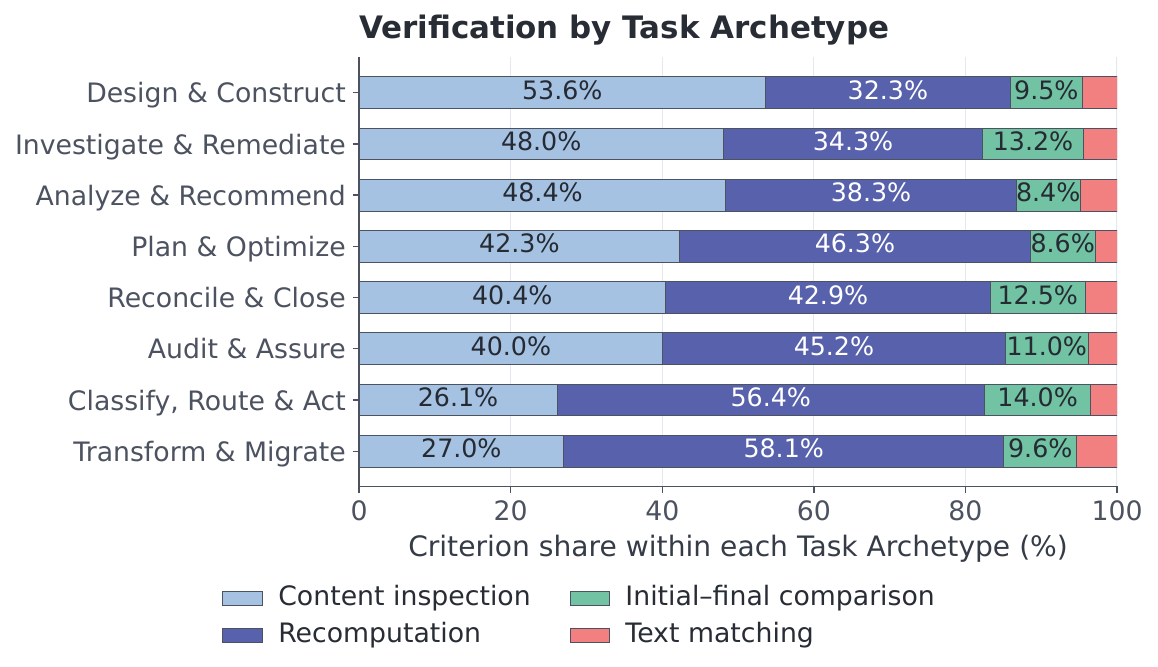}
\caption{\textbf{Verification-method composition varies across task archetypes.}
Each row divides method counts by that archetype's total criterion count
and sums to 100\%; criteria are unweighted by rubric scores.}
\label{fig:archetype-verification-stacked}
\end{figure}

\paragraph{Different compositions under a shared rubric format.}
Figure~\ref{fig:archetype-verification-stacked} normalizes criterion
counts within each archetype. Content inspection accounts for
53.6\% of Design \& Construct criteria, while recomputation
accounts for 56.4\% of Classify, Route \& Act and 58.1\% of
Transform \& Migrate criteria. Plan \& Optimize also devotes
46.3\% of its criteria to recomputation.
The archetypes therefore share a verification vocabulary while
allocating checks differently. This pattern is consistent with
compiling task-specific obligations under a common completion
representation.

\section{Task Structure and Execution Length}\label{app:task-horizon}

\begin{figure}[htbp]
\centering
\includegraphics[width=\linewidth]{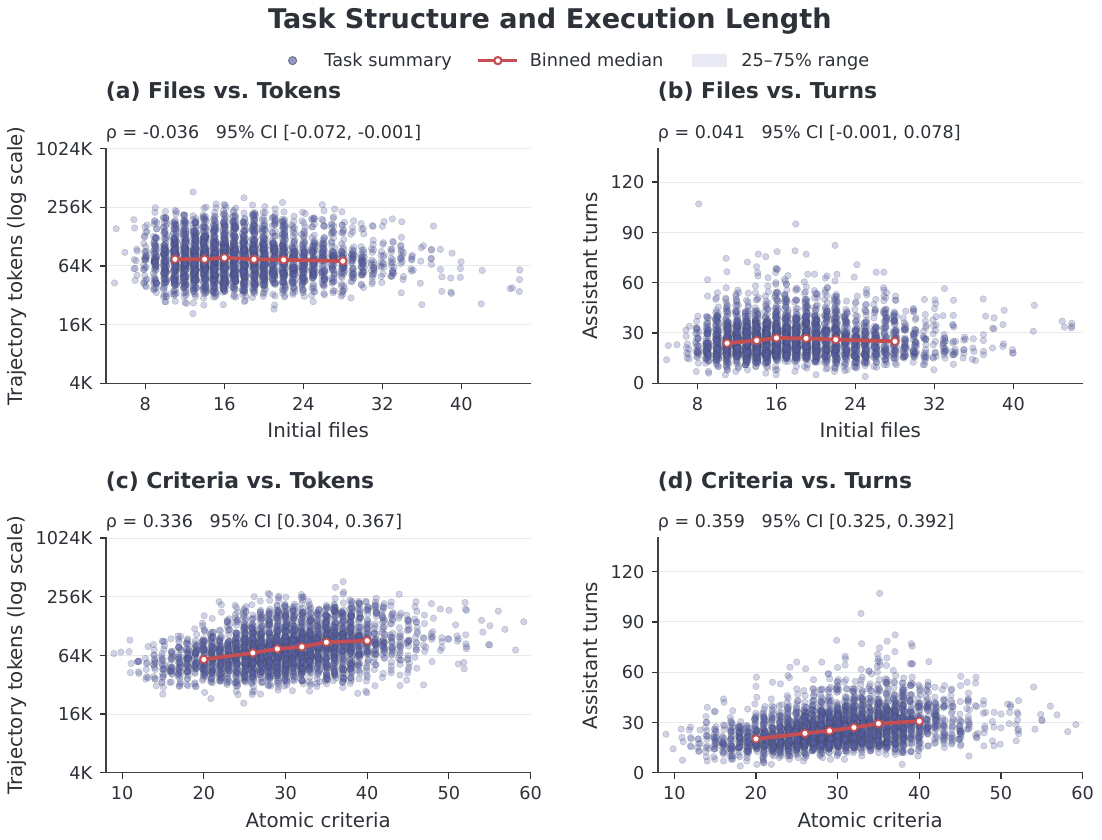}
\caption{\textbf{More atomic outcome criteria are associated with longer executions.}
Points summarize the 3,278 retained tasks by averaging the available
solver-specific medians over repeated runs; horizontal jitter aids display.
Red curves show binned medians; shaded bands show the 25th--75th
percentiles of task summaries.
Annotations report Spearman correlations and 95\% confidence
intervals from bootstrap resampling clustered by source task.
Token axes are logarithmic; $\mathrm{K}=1{,}024$.}
\label{fig:task-horizon-overall}
\end{figure}

Atomic outcome criteria are positively associated with both measures
of execution length in Figure~\ref{fig:task-horizon-overall}.
Their Spearman correlations are $\rho=0.336$ with recorded trajectory
tokens and $\rho=0.359$ with assistant turns, with 95\% confidence
intervals of $[0.304,0.367]$ and $[0.325,0.392]$, respectively.
The binned medians rise with criterion count for both measures.
Initial file counts have correlations close to zero
($\rho=-0.036$ and $\rho=0.041$).

This pattern aligns with \Interlock{}'s emphasis on preserving the
requirements that define a complete outcome. The rubric expresses
conditions on evidence-supported decisions, jointly constrained
components, and required downstream effects. Grounding connects
decisions to their controlling sources, Coupling requires their
uses to remain jointly valid, and Realization specifies the effects
that must be delivered. The positive criterion-count associations
connect this explicit outcome specification to the sustained
execution observed in the generated tasks.

\paragraph{Task-level aggregation.}
The analysis uses the 32,370 normally terminated rollouts on the
3,278 retained tasks
and the token-counting rules in Appendix~\ref{app:trajectory-statistics}.
For each task, we take the median over repeated runs within
each solver and then the arithmetic mean of the available
solver-specific medians. Each task instance receives equal
weight in the correlations. Confidence intervals use 2,000
bootstrap resamples clustered by source task pack, retaining
all associated instances together.

\clearpage

\clearpage
\section{Further Case Studies of Synthesized Tasks}\label{app:synthesized-case-studies}

These cases extend Figure~\ref{fig:synthesized-cases-main} to scientific
protocols, infrastructure planning, and payroll records.
Each pair compares two rollout attempts of the same Qwen backbone
with identical public requests, initial workspaces, and rubrics.
The field-trial case uses Qwen3.6-35B-A3B; the other four cases use
Qwen3.6-27B. Attempt 1 and attempt 2 identify the displayed pair of runs.
The diagrams reconstruct requirements from public sources and summarize
recorded tool events and final artifacts. Counts denote passed/total
binary criteria in the saved Judge reports; independent file audits
are stated separately. These comparisons characterize variation across
attempts on a task, while Table~\ref{tab:results} provides the aggregate
training comparison. The outcomes are plans and records delivered
in task workspaces.

\subsection{Field-Trial Protocol: Decisions Must Remain Executable and Interpretable}
\label{app:case-field-trial}

The research lead needs a 2027 field-trial authorization protocol and
a consistent reviewer note. In attempt 1, Qwen3.6-35B-A3B reverses the public risk
condition: the confidence-interval branch flags risk when it excludes
a yield loss above two bushels per acre, whereas the rule flags risk
when it cannot exclude that loss. The plan also retains conditional
AP7 treatment at SD-L2 despite its conflict with the associated
May 9 pre-emerge constraint. This constraint applies to that operation
and AP7 feasibility. Seasonal workload totals do not supply the
required dated allocation of observations and staff. The protocol
further makes grower-paid resprays reimbursable, and its secondary
weed endpoint becomes primary in the reviewer note.

\begin{figure}[htbp]
\centering
\includegraphics[width=\linewidth]{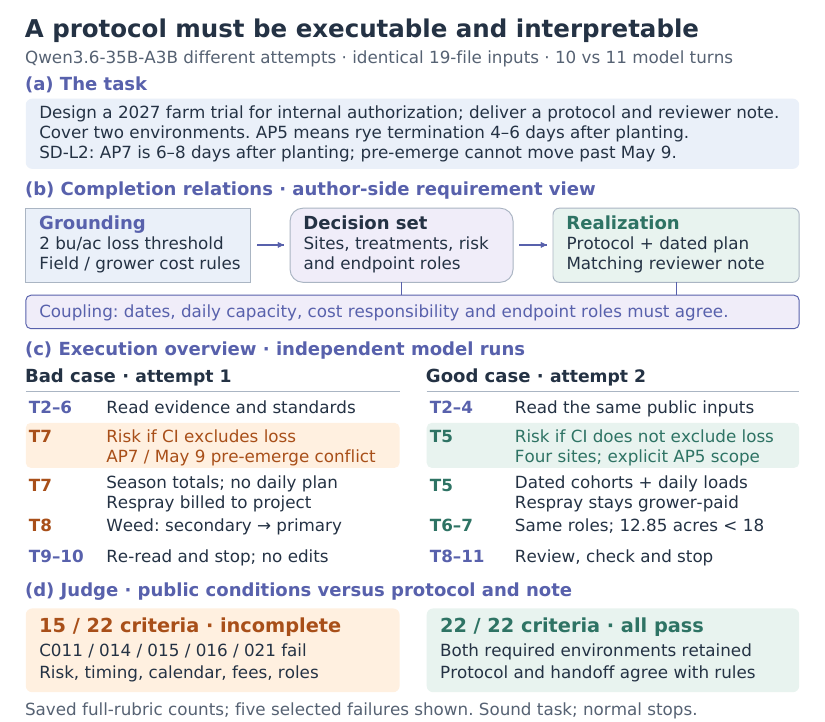}
\caption{\textbf{A protocol must preserve both feasibility and interpretation.}
Five selected failures connect the risk rule, treatment timing, daily
resources, payer boundary, and endpoint roles. AP5 and AP7 mean
termination 4--6 and 6--8 days after planting, respectively; the risk
threshold is two bushels per acre. Full saved Judge counts are 15/22
and 22/22 for attempts 1 and 2 of Qwen3.6-35B-A3B, respectively;
$T$ denotes model turns.}
\label{fig:case-field-trial}
\end{figure}

Attempt 2 of the same Qwen3.6-35B-A3B model uses four sites with AP5, preserves both
required environments, and meets the public minimum of two sites per
environment. It retains the risk condition, dated activities, payer
boundary, and endpoint roles. The saved Judge counts for attempts 1
and 2 are 15/22 and 22/22. The five focal failures in Figure~\ref{fig:case-field-trial}
connect grounded interpretation to a jointly feasible design and to
the measurements and explanations that must realize it.

\clearpage
\subsection{Charging Networks: Capacity Must Stay Attached to Its Resource}
\label{app:case-ev-network}

The campus charging plan couples complete demand dispositions with
energy delivery, equipment ratings, and nested electrical limits.
In attempt 1, Qwen3.6-27B correctly computes 53 kW of headroom for panel PA-1,
then assigns 88.5 kW to that panel by using the combined headroom of
PA-1 and another panel. It also omits nine demand sources, assigns
the protected 144 kWh FLT-207 demand to a single 9.6 kW port that
can supply at most $9.6\times4.5=43.2$ kWh within its service window,
and records equipment nameplates inconsistent with the selected
models and quantities. These choices remain in the final files
after syntax repairs and a concluding check.

\begin{figure}[htbp]
\centering
\includegraphics[width=\linewidth]{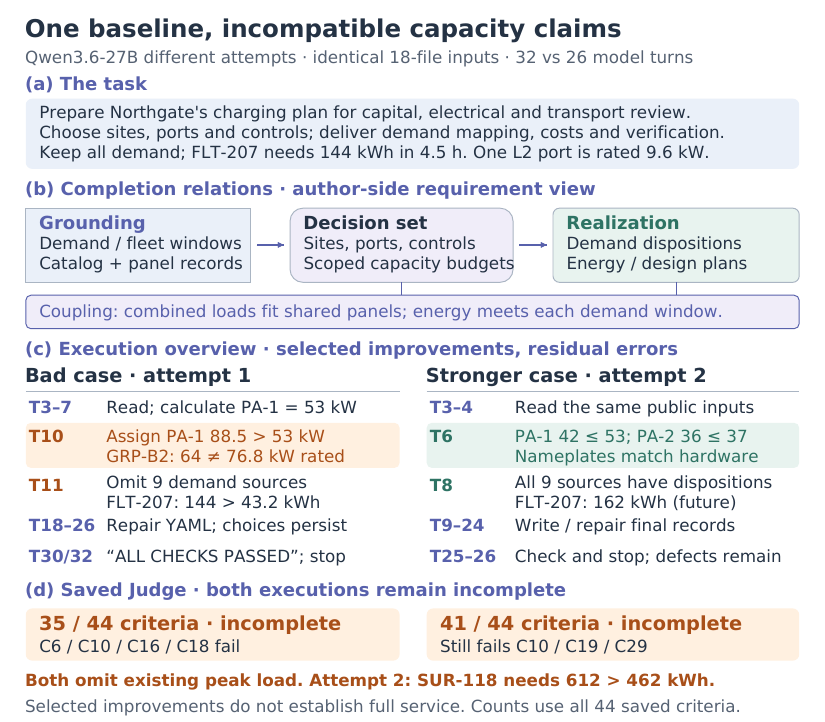}
\caption{\textbf{Correct capacity calculations must constrain the final allocation.}
The diagram traces demand coverage, panel scope, energy-service windows,
and hardware nameplates. Attempt 2 meets the highlighted object constraints but
retains an energy deficit, missing existing peak loads, and a cost
inconsistency. Both Qwen3.6-27B attempts remain incomplete, at 35/44
and 41/44 criteria; the FLT-207 counterpart is marked as future service.}
\label{fig:case-ev-network}
\end{figure}

Attempt 2 of the same Qwen3.6-27B model preserves the nine demand dispositions and the focal panel
and nameplate constraints. Its 162 kWh FLT-207 allocation is explicitly
a future design. A different demand, SUR-118, still requires 612 kWh
against 462 kWh of planned capacity, and both executions omit existing
loads from the campus peak. Attempt 2 also retains a cost-summary
mismatch. The saved Judge counts for attempts 1 and 2 are 35/44 and
41/44, with both runs
incomplete. Figure~\ref{fig:case-ev-network} locates the gap between a
correct local calculation and its use under the proper resource scope,
while preserving the outstanding joint obligations in each plan.

\clearpage
\subsection{Payroll Close: Amounts, Authority, and Accounting Effects}
\label{app:case-payroll-close}

The payroll close requires six linked deliverables for Payroll,
Treasury, Accounting, and Employee Relations. In attempt 1, Qwen3.6-27B records
E1004's pending exposure as \$34.50 instead of the source-supported
\$448.50, marks E1011's \$35 refund as authorized using a time-correction
approval, and omits E1005's required \$1,760 gross-cost row while
holding the \$1,296.80 net payment. Attempt 2 of the same model
handles those focal obligations, including a recorded addition of
the gross-cost row at turn 19 without releasing the held cash.
The case connects source scope and action-specific authority to
distinct payment and accounting effects.

\begin{figure}[htbp]
\centering
\includegraphics[width=\linewidth]{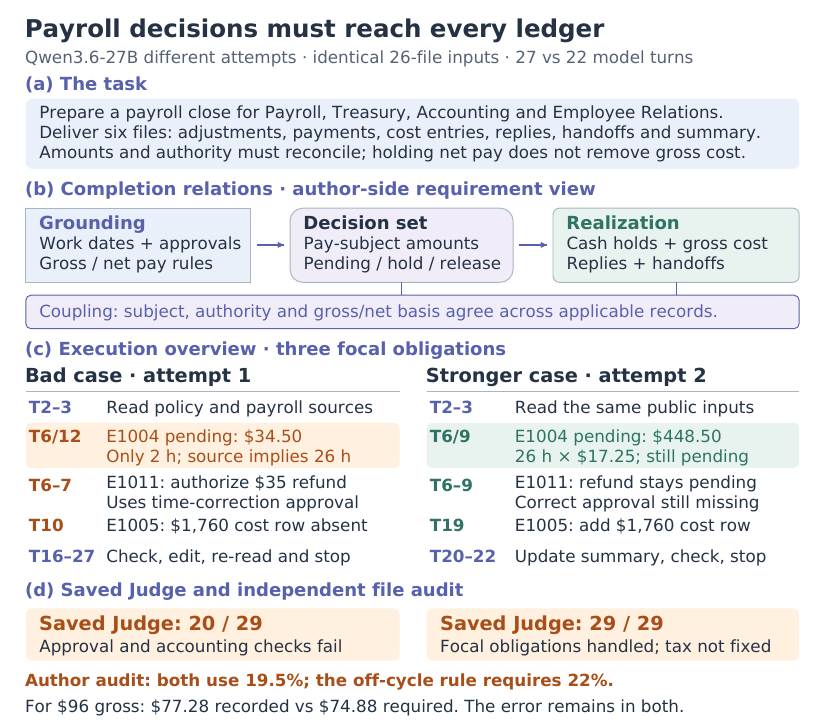}
\caption{\textbf{Cash holds still require corresponding cost records.}
The three focal obligations concern E1004's pending amount, E1011's
refund authority, and E1005's gross-cost entry. The \$1,296.80 cash
hold and \$1,760 gross cost refer to different required records.
Saved Judge counts for the two Qwen3.6-27B attempts are 20/29 and
29/29; the independent audit retains
the off-cycle tax error shared by both executions.}
\label{fig:case-payroll-close}
\end{figure}

The saved Judge reports 20/29 and 29/29 for attempts 1 and 2, while the independent file
audit finds a shared off-cycle tax error. Both executions use 19.5\%
where the public rule requires 22\%; for \$96 gross, they record
\$77.28 net instead of \$74.88. Figure~\ref{fig:case-payroll-close}
therefore reports the saved judgments alongside this remaining
artifact-level error. Handling the three displayed obligations and
preserving the correct tax basis are separate completion requirements.

\clearpage
\section{Algorithm and Experimental Details}\label{app:implementation-details}

\subsection{Task Synthesis and Training Algorithm}\label{app:algorithm}

Algorithm~\ref{alg:interlock} summarizes task construction, admission,
and completion-filtered training.

\begin{algorithm}[t]
\caption{\Interlock{}: task synthesis and completion-filtered training.}
\label{alg:interlock}
\small
\begin{algorithmic}[1]

\Require Blueprints $\mathcal B$, validation policy $\pi_v$,
         backbones $\{\pi_0^{(k)}\}$, attempts $M$,
         repair budget $R$

\Ensure Admitted tasks $\mathcal P$,
        training sets $\{\mathcal D_{\mathrm{RFT}}^{(k)}\}$,
        trained policies $\{\pi_{\theta_k}\}$

\State $\mathcal P\gets\varnothing$

\ForAll{$b\in\mathcal B$}
    \State $\mathcal G_b\gets
        \Call{BuildCompletionGraph}{b}$
    \State $c\gets
        \Call{CompileTaskPack}{b,\mathcal G_b}$
        \Comment{Eq.~\ref{eq:task-construction}}
    \State $x\gets
        \Call{ReviewAndAdmit}{c,\pi_v,R}$
    \If{$x\neq\varnothing$}
        \State $\mathcal P\gets\mathcal P\cup\{x\}$
    \EndIf
\EndFor

\ForAll{backbones $k$}
    \State $\mathcal T_{\mathrm{ok}}^{(k)}\gets
        \Call{CollectAndJudge}{\pi_0^{(k)},\mathcal P,M}$
    \State $\mathcal D_{\mathrm{RFT}}^{(k)}\gets
        \{(x,\tau)\in\mathcal T_{\mathrm{ok}}^{(k)}:
        A_V(x,\tau)=1\}$
    \State $\pi_{\theta_k}\gets
        \Call{SFT}{\pi_0^{(k)},
        \mathcal D_{\mathrm{RFT}}^{(k)}}$
\EndFor

\State \Return
    $\mathcal P,
    \{\mathcal D_{\mathrm{RFT}}^{(k)}\},
    \{\pi_{\theta_k}\}$

\end{algorithmic}
\end{algorithm}

\clearpage
\subsection{Detailed Experimental Setup}\label{app:experimental-setup}

\paragraph{Models and training conditions.}
We evaluate \Interlock{} on two backbones, \textbf{Qwen3.6-27B} and \textbf{Qwen3.6-35B-A3B}. For each backbone, the main experiment compares its original checkpoint, a model fine-tuned on released ClawGym trajectories, and a model trained with \Interlock{} through rejection-sampling fine-tuning (RFT). All three variants are evaluated on AutomationBench, JobBench, and Harvey LAB. The comparison examines the training value of the generated experience and its transfer to external task distributions.

Appendix~\ref{app:task-statistics} reports statistics of the retained
task corpus, including archetype coverage, workspace and rubric size,
and verification methods.

\textbf{Original.} The original checkpoint is the starting policy before either data-based training procedure. It provides the reference for measuring the effect of additional training within each backbone.

\textbf{ClawGym-SFT.} We directly use the trajectories released by ClawGym~\citep{bai2026clawgym} for supervised fine-tuning of each starting model. This baseline represents training on an existing source of environment-grounded agent demonstrations.

\textbf{Interlock-RFT.} In the proposed protocol, each backbone independently executes \Interlock{} tasks and supplies its own rollout pool. Normally ended trajectories satisfying the acceptance rule in Section~\ref{learning-from-scored-rollouts} are used for SFT initialized from the same backbone. Rollout generation, acceptance, and parameter updates remain separate for Qwen3.6-27B and Qwen3.6-35B-A3B. RFT denotes this combination of rejection sampling and SFT.

\paragraph{Task screening and rollout accounting.}
The reported corpus contains 3,278 synthesized tasks retained after
screening. Each retained task has at least one accepted
rollout from Qwen3.6-27B and at least one from
Qwen3.6-35B-A3B. This criterion defines the task set used in the
reported experiments.
For each retained task, each backbone makes $M=5$ rollout attempts
from reset workspaces, giving $3{,}278\times2\times5=32{,}780$ attempts,
or 16,390 per backbone. We exclude 410 attempts that do not terminate
normally because of execution anomalies or incomplete execution,
leaving 32,370 normally terminated trajectories for trajectory analysis.
Normal termination concerns execution status; trajectory acceptance is
determined separately by $A_V(x,\tau)$.

From its own normally terminated rollout pool, Qwen3.6-27B selects
8,957 accepted trajectories for its RFT training set.
Qwen3.6-35B-A3B independently selects 8,957 accepted trajectories
from its own pool for its own training. The two training sets contain
17,914 trajectories in total, with approximately $154\mathrm{M}$
supervised tokens per backbone ($\mathrm{M}=10^6$).
Supervised tokens are assistant-generated positions that contribute
to the training loss, including reasoning, responses, and tool calls.
System and user messages, tool-returned observations, and tool schemas
provide conditioning context and are excluded from this supervision count.

These conditions compare the full \Interlock{} data-generation and training procedure with both the original model and a released-trajectory SFT alternative. The ClawGym baseline uses its published demonstrations, whereas \Interlock{} generates and selects demonstrations from the target backbone itself. Training-data sizes and optimization budgets are specified separately for each condition.

\textbf{Implementation details.} Qwen3.6-27B and Qwen3.6-35B-A3B
use the same SFT configuration in Table~\ref{tab:sft-configuration}.
Each model is trained for five epochs on 16 NVIDIA H200 GPUs with
141\,GB of memory per GPU. Both backbones use thinking mode during
training and evaluation, and reasoning tokens contribute to the
supervised loss. The approximately $154\mathrm{M}$ supervised tokens
per backbone describe the selected training data before epoch repetition.

\begin{table}[htbp]
\centering\small
\caption{Shared SFT configuration for Qwen3.6-27B and Qwen3.6-35B-A3B.}
\label{tab:sft-configuration}
\begin{tabular}{@{}ll@{}}
\toprule
\textbf{Parameter} & \textbf{Value} \\
\midrule
Maximum sequence length & $200\mathrm{k}$ tokens \\
Learning rate & $2\times10^{-5}$ \\
Batch size & 64 \\
Training epochs & 5 \\
Warmup ratio & 0.03 \\
Learning-rate schedule & Cosine \\
Training hardware & $16\times$ NVIDIA H200 (141\,GB each) \\
\bottomrule
\end{tabular}
\end{table}

\paragraph{Benchmarks and evaluation.}

We assess out-of-domain (OOD) generalization by transferring from the \Interlock{} synthetic task distribution to three independently constructed benchmark distributions. We use the public releases of AutomationBench and Harvey LAB. The evaluated benchmark instances are excluded from \Interlock{}'s synthesis inputs and RFT data. This setting tests whether experience acquired in generated task environments supports completion of externally specified tasks.

\textbf{AutomationBench} contains 600 public benchmark tasks, with 100 tasks in each of six business domains. Agents coordinate actions across applications through APIs, and programmatic assertions evaluate the final application state. We report \emph{pass rate}: the percentage of task executions satisfying every required assertion~\citep{shepard2026automationbench}.

\textbf{JobBench} contains 130 tasks across 35 occupations, divided into \emph{Main} and \emph{Easy} sets of 65 tasks each. The tasks require agents to reconcile heterogeneous reference materials and produce professional deliverables. Main includes more demanding reasoning and source reconciliation; Easy uses self-contained reference workspaces without web search. We run JobBench with the \textbf{OpenCode} harness and use \textbf{GPT-5.5} to return an overall binary pass/fail judgment for each weighted rubric group~\citep{li2026jobbench}.

\textbf{Harvey LAB} contains 1,251 tasks across 24 legal practice areas in the public release used in our evaluation. Tasks combine a client matter, supporting documents, and reviewable work products. We use \textbf{GPT-5.5} as the judge for its deliverable-specific binary criteria, which assess factual support, legal analysis, and required content~\citep{harveylab2026}.

\textbf{Reported metrics.} The main table reports AutomationBench task pass rate, JobBench-Main weighted score, JobBench-Easy weighted score, and Harvey LAB criterion pass rate. AutomationBench averages binary task-pass decisions. Harvey LAB criterion pass rate pools the number of passed binary criterion verdicts over the total number evaluated across the corresponding tasks and runs.

The JobBench evaluator defines a task's weighted score as \texttt{total\_score}/\texttt{max\_score}. With valid judgments for every rubric group, we report the task-averaged weighted score
\begin{equation}
\mathrm{Score}_{\mathrm{JobBench}}
=\frac{100}{N}\sum_{i=1}^{N}
\frac{\sum_j w_{ij}b_{ij}}{\sum_j w_{ij}}.
\label{eq:jobbench-weighted-score}
\end{equation}
Here, $N$ is the number of evaluated tasks in the corresponding Main or Easy split, $j$ indexes rubric groups within task $i$, $w_{ij}$ is the rubric weight, and $b_{ij}\in\{0,1\}$ is the Judge's overall pass indicator for that rubric group. Each task is normalized by its own total rubric weight before the task scores are averaged equally.

All metrics use a 0--100 scale, with higher values indicating better performance. JobBench improvements are reported in score points; AutomationBench and Harvey LAB rate improvements are reported in percentage points. Repeated executions and judge assessments follow the same schedule for all compared conditions. Evaluation uses each benchmark's supplied assertions or rubrics.

\textbf{Evaluation configuration.} Comparisons are organized within a common benchmark version and task split. JobBench uses the OpenCode harness, and both JobBench and Harvey LAB use GPT-5.5 as the judge.

\clearpage
\section{Prompts and Contracts for Graph-Grounded Synthesis}
\label{app:interlock-prompts}

\Interlock{} constructs tasks from a frozen source specification that retains
the full task requirements. Its completion graph constrains workspace construction, task
writing, and rubric compilation; separate source and public-input reviews
check requirement preservation and public support. This appendix presents
six selected prompt excerpts and four compact data-contract summaries.
The excerpts are condensed English translations of the current
implementation's Chinese templates. Repeated self-checks and runtime
boilerplate are omitted. Field names and role boundaries follow the code.

\subsection{Frozen Source and Completion Graph}
\label{app:prompt-graph-contract}

Offline specification authoring declares explicit evidence, decision,
and effect nodes, together with relation endpoints and joint conditions.
Co-occurring roles or shared pivot names do not imply edges.
The selected contract is frozen before the production roles run.
The source snapshot retains the full task description and its intended use,
all required business clauses, archetype requirements, and any selected
artifact obligations or motif. This source remains available for review
even when a generated graph or artifact omits an obligation.

\begin{InterlockPromptBox}{Contract 1: Frozen source and typed graph}
source_spec.json
  schema_version: "completion_source.v1"
  request
  brief
  source_requirements: [{id, kind, content}]
  completion_contract

completion_graph.json
  schema_version: "completion_graph.v1"
  source: {task_id, domain_id, neighborhood_id,
           task_archetype_id}
  evidence: [...]
  decisions: [...]
  effects: [...]
  grounding: [...]
  coupling: [...]
  realization: [...]

Common fields of every node and relation:
  id
  source_requirement_ids: [source requirement IDs]
  required_stages: [workspace | task | rubric]
\end{InterlockPromptBox}

The records above summarize fields rather than reproduce the full JSON
Schema. Source IDs distinguish the overall task request
(\InterlockField{commission:}), its business requirements
(\InterlockField{requirement:}), and archetype requirements
(\InterlockField{archetype:}); declared artifact obligations and motifs
retain their own source entries. Each entry preserves its original content.

\begin{center}
\small
\renewcommand{\arraystretch}{1.13}
\begin{tabularx}{\linewidth}{@{}lX@{}}
\toprule
Element & Fields in addition to the common fields \\
\midrule
Evidence $E$ & \InterlockField{artifact_role_id}. \\
Decision $D$ & \InterlockField{description}, and exactly one of
  \InterlockField{pivot_id} or \InterlockField{state_id}. \\
Effect $Y$ & \InterlockField{downstream_role_id} and the four fields below. \\
Grounding $G$ & \InterlockField{evidence_id}, \InterlockField{decision_id}. \\
Coupling $C$ & \InterlockField{member_ids}, \InterlockField{condition},
  \InterlockField{public_basis_role_ids}. Members are decisions or effects. \\
Realization $R$ & \InterlockField{decision_id}, \InterlockField{effect_id}. \\
\bottomrule
\end{tabularx}
\end{center}

\begin{InterlockPromptBox}{Effect fields: public basis, population, and conditions}
B      = public_basis_role_ids
Omega  = population {
           business_object_ids,
           source_role_ids,
           description
         }
alpha  = applicability
phi    = outcome_condition
\end{InterlockPromptBox}

The effect fields store source references and natural-language conditions.
The source population is established from public inputs; applicability
and correct realizations remain to be determined from those inputs.
These declarations do not contain answer values or executable graders.
Python checks reference integrity and stage obligations: each relation
requires rubric coverage and at least one public-input stage
(\InterlockField{workspace} or \InterlockField{task}), and each effect
requires rubric coverage. Coupling relations have at least two members
when declared. The graph is a requirements representation, not an
execution-order DAG.

\subsection{Selected Construction and Review Prompts}
\label{app:prompt-excerpts}

The shared graph instructions are inserted into each production prompt.
The builder's and setter's archetype guidance comes from the frozen
brief; contract fields and the rubric coverage floor are supplied from
their corresponding schemas and declarations.
The source judge sees the frozen specification, graph, actual artifacts,
and carrier maps. The public critic sees only the public task and
workspace plus the rubric under review.

\paragraph{Shared graph contract.}
This block keeps the same requirements in view throughout construction
and makes the carrier mappings explicit.
\label{app:prompt-shared-graph}
\begin{InterlockPromptBox}{Prompt 1: Shared completion-graph instructions}
Read the frozen source specification and completion graph. All three production stages use the same version. Do not alter its requirements, nodes, relations, conditions, or required stages. The graph specifies business relations that must hold, not the solver's execution order.

Grounding preserves the evidence and rules supporting decisions. Coupling preserves joint conditions over its members. Realization preserves the downstream effects of decisions. Preserve each effect's public basis, source population, applicability, and outcome condition. Determine the population from public sources, not from the solver's eventual submission.

Record actual carriers using graph IDs and file locations; rubric carriers also identify criteria. A mapping locates content and does not establish semantic preservation. Keep the graph and mappings outside the solver workspace.

Before finishing, check preservation, actual carrier locations, and the write boundary of your role.
\end{InterlockPromptBox}

\paragraph{Workspace construction.}
The builder materializes evidence while leaving the controlling decisions
and complete outcomes for the solver.
\label{app:prompt-world-builder}
\begin{InterlockPromptBox}{Prompt 2: World builder}
Build the workspace for the full task. The task statement and rubric are written by later roles and are unavailable to you. Provide the evidence, professional methods, existing records, resources, authorization, and interacting constraints needed across the full scope of the work.

Preserve independently identifiable sources, original conflicts, relevant objects and branches, and nontrivial derivations. Do not pre-join evidence, resolve disputes, preselect objects whose applicability remains to be determined, or provide completed deliverables. Organize files by business purpose; file count is not a complexity target.

Write only workspace/ and a private, value-free evidence_index.json. Its completion_carriers locate facts, rules, source populations, and existing states in the actual workspace. Controlling decisions remain unresolved.
\end{InterlockPromptBox}

\paragraph{Public task specification.}
The setter turns the full task requirements into publicly supported
obligations while preserving freedom over the solution procedure.
\label{app:prompt-task-setter}
\begin{InterlockPromptBox}{Prompt 3: Task setter}
Read the workspace, evidence index, task brief, frozen source specification, and completion graph. Express every necessary obligation as a public requirement or an invariant needed to achieve the public goal.

Write a task statement identifying the intended users, purpose, deliverables, constraints, and authorization boundaries. Preserve necessary decisions, record modifications, conditional obligations, and relationships between deliverables. State outcomes without prescribing the reading, calculation, or writing order. Do not disclose computed results, selected objects, or resolved conflicts.

Write the deliverable contract and its completion_carriers. Every contractual obligation must have a basis in task.md or the public workspace. The private contract cannot introduce additional acceptance requirements or sending authorization. Modify only task.md and deliverable_contract.json.
\end{InterlockPromptBox}

\paragraph{Outcome rubric compilation.}
The compiler links required decisions, joint conditions, and effects to
criteria that can be judged from public evidence and delivered artifacts.
\label{app:prompt-rubric-compiler}
\begin{InterlockPromptBox}{Prompt 4: Rubric compiler}
Compile acceptance criteria from the public task and workspace, using the frozen graph and private indices to check coverage. Do not modify public inputs or generate a reference answer.

Each criterion tests one proposition with a binary result. Identify its final-artifact target, concrete public basis, and appropriate verification method. Preserve the full object population, applicability conditions, exceptions, and permitted equivalent expressions. Recomputed criteria describe a derivation in prose without answer values. Cross-artifact consistency must depend on checks of source correctness.

Write rubric_spec.json and a separate rubric_carriers.json. Every required effect must map to substantive outcome criteria, and joint constraints must remain jointly checked. Keep private graph identifiers in the carrier map. Unpublished source obligations cannot directly penalize the solver.
\end{InterlockPromptBox}

\paragraph{Independent source-coverage judgment.}
This review starts from the original task specification, including requirements
that might be jointly omitted from the graph, request, and rubric.
\label{app:prompt-source-judge}
\begin{InterlockPromptBox}{Prompt 5: Source-coverage judge}
Independently review preservation from the original task specification to the current task pack. Read the frozen specification, graph, actual stage artifacts, carrier maps, expected review targets, and prior source findings. Do not read solver trajectories or final outputs, public critic conclusions, or solver scores.

Review every source requirement, including obligations jointly omitted from the graph and generated artifacts. Then judge every expected relation-stage pair from its actual carrier content. File existence, matching IDs, and producer declarations do not establish preservation.

Check Grounding evidence and controlling rules, Coupling members and joint conditions, and Realization basis, source population, applicability, and outcome conditions. Do not let an omitted required stage or a reduced population conceal an original obligation.

Return preserved, defective, or undetermined with evidence for each item. Every non-preserved judgment needs an unresolved finding. Explicitly reassess every historical finding, including valid findings retained from interrupted calls. Identify the responsible producer without changing inputs, writing answers, or predicting solver success.

Write only source_review.json. Code aggregates the overall status.
\end{InterlockPromptBox}

\paragraph{Independent public-input criticism.}
This complementary review checks the task as supported by the materials
available to the solver, without access to the source specification.
\label{app:prompt-rubric-critic}
\begin{InterlockPromptBox}{Prompt 6: Public-input rubric critic}
Review only task.md, the public workspace, and the proposed rubric_spec.json. The task brief, private contracts, evidence index, author notes, prior reviews, and solver trajectory are unavailable. Do not use assumptions outside public inputs to supply missing acceptance grounds.

Recover the complete public obligations, then check rubric coverage. Verify object scope, conditions, exceptions, authorization, joint consistency, and public decidability. Distinguish legitimate unresolved cases and equivalent expressions from missing work. Report duplicated predicates only when their object and condition scopes actually overlap.

Write only rubric_critique.json, with evidence and repair guidance for the responsible producer. Do not edit the rubric, construct reference deliverables, prove solvability, or predict difficulty. Return clean with no findings when no supported defect is found.
\end{InterlockPromptBox}

\subsection{Carrier and Rubric Contracts}
\label{app:prompt-carrier-rubric}

Each producer records where its graph requirements are carried.
The same mapping structure appears in the evidence index
($\mu_W$), deliverable contract ($\mu_T$), and separate rubric
carrier file ($\mu_V$).
One graph element may have several carriers, and one criterion may
serve several graph elements.

\begin{InterlockPromptBox}{Contract 2: Stage-specific carrier mappings}
completion_carriers: [
  {
    graph_id,
    carriers: [
      {path, selector, criterion_id?}
    ]
  }
]

mu_W: evidence_index.json
mu_T: deliverable_contract.json
mu_V: rubric_carriers.json
\end{InterlockPromptBox}

Question marks mark optional fields in these summaries.
For a carrier pointing to \InterlockField{rubric_spec.json},
\InterlockField{criterion_id} is required and must name an existing
criterion. Each graph element must be covered at its declared stages;
each required rubric mapping must identify an actual criterion.
File and reference checks are performed by code. The source judge
then inspects the content to assess preservation.
Future solver outputs need not exist in the initial workspace.

\begin{InterlockPromptBox}{Contract 3: Compact outcome rubric}
schema_version: "rubric_spec.v6"
quality_floor
rubrics: [
  {
    id, title, weight, consequence,
    criteria: [
      {
        id, criterion,
        target: {path, selector},
        public_basis: [{path, selector}],
        verification_method,
        importance, credit_mode,
        derivation?, depends_on?
      }
    ]
  }
]

verification_method:
  recomputed | initial_final_diff | text_match | read
importance: essential | supporting
credit_mode: change | preserve
consequence: critical | forbidden | material | supporting
\end{InterlockPromptBox}

\InterlockField{criterion} states a binary satisfaction condition;
\InterlockField{target} locates the delivered content to inspect;
\InterlockField{public_basis} identifies supporting locations in
\InterlockField{task.md} or \InterlockField{workspace/}.
\InterlockField{derivation} is a prose recomputation rule required for
\InterlockField{recomputed} criteria and absent for other methods.
\InterlockField{depends_on} references criterion dependencies for
scoring; it is distinct from the completion graph.
The compiler writes the compact specification, and code renders
\InterlockField{rubric.md}. Private graph mappings remain in
\InterlockField{rubric_carriers.json}.

Required criteria express mandatory public obligations, including their
applicability conditions and permitted alternatives. An accepted trajectory
satisfies $A_V(x,\tau)=1$: every required criterion passes and any separately
declared quality threshold is met. Mandatory obligations and quality
preferences are distinguished by their public meaning, as described in
Appendix~\ref{app:outcome-scoring}. Weights, importance labels, credit modes,
and criterion dependencies support the diagnostic outcome scores in
Appendix~\ref{app:diagnostic-scoring}; they do not waive mandatory obligations.

\subsection{Review Results and Task Admission}
\label{app:prompt-review-contract}

The source judge reports individual judgments rather than an aggregate
pass declaration. Code constructs the expected source-requirement and
relation--stage sets from the frozen inputs and checks that each target
has exactly one judgment. Previous findings also require explicit
dispositions.

\begin{InterlockPromptBox}{Contract 4: Independent source-review report}
schema_version: "source_review.v1"
basis: [evidence descriptions]
requirements: [
  {requirement_id, status, evidence}
]
relations: [
  {relation_id, stage, status, evidence}
]
findings: [
  {
    finding_id, status, owner_stage,
    requirement_ids, relation_ids, stages,
    problem, evidence, reproduction, suggested_change
  }
]
reviewed_findings: [
  {finding_id, disposition, evidence}
]

Preservation status: preserved | defective | undetermined
New finding status: confirmed | undetermined
Prior finding disposition: confirmed | dismissed | undetermined
owner_stage:
  source_spec | completion_graph | world_builder |
  task_setter | rubric_compiler
\end{InterlockPromptBox}

Code aggregates the source result as passed, defective, or undetermined.
An unresolved item requires an associated finding; silence in a later
report does not close an earlier one.
Source findings and public-input findings are kept in separate records
and adjudicated using their respective evidence.
The post-execution judge evaluates the public task against initial and
final files and their differences, without seeing the source graph,
carrier maps, source findings, or solver trajectory.

Task admission requires both a passed source-preservation review and
a clean public-input critic report, a normally terminated real execution,
complete criterion judgments, and a sound package-quality assessment
with no outstanding confirmed or undetermined defects.
The two reviews establish source coverage and public support;
the post-execution judge separately assesses the resulting artifacts
and package quality.
Trajectory acceptance is recorded separately as $A_V(x,\tau)=1$
and is not required for task admission.
Confirmed defects in materials, the request, or the rubric return to
their responsible generation stage. Source-specification and graph
defects require offline correction and keep the candidate internal.
Revisions receive fresh reviews. The two review views share at most one
generation-time rubric recompilation; post-execution revision follows
the separately bounded quality-repair budget.


\section{Benchmarks for Long-Horizon Agents}\label{app:related-benchmarks}

Long-horizon benchmarks provide concrete settings in which successful interaction must produce the requested outcome.
WebArena~\citep{zhou2023webarena} and OSWorld~\citep{xie2024osworld} evaluate realistic web and desktop tasks with execution-based outcome checks.
AppWorld~\citep{trivedi2024appworld} tests stateful multi-application coding and unintended changes, and ToolSandbox~\citep{lu2025toolsandbox} emphasizes implicit state dependencies and intermediate milestones.
These settings connect task success to environmental state and applicable rules, motivating our treatment of completion as a set of linked requirements.

WorkArena++~\citep{boisvert2024workarenapp} targets compositional planning and reasoning in knowledge work, while TheAgentCompany~\citep{xu2024theagentcompany} combines websites, code execution, and workplace communication.
Terminal-Bench~\citep{merrill2026terminalbench} couples challenging command-line tasks with dedicated environments and tests, and Long-Horizon-Terminal-Bench~\citep{li2026longhorizonterminalbench} evaluates extended workflows with graded subtasks and partial credit.
Their emphasis on coordinated work and outcome evaluation complements \Interlock{}'s focus on preserving completion requirements during task synthesis.

Our external evaluations span three forms of consequential work.
AutomationBench~\citep{shepard2026automationbench} brings together cross-application orchestration, autonomous API discovery, and layered policy compliance.
JobBench~\citep{li2026jobbench} evaluates expert-prioritized delegations in heterogeneous reference workspaces using fact-anchored chains of binary rubric criteria.
Harvey LAB~\citep{harveylab2026} evaluates legal work in document-rich matter workspaces through deliverable-specific binary criteria covering factual support, analysis, and required content.
Using each benchmark's own evaluation protocol, these independently authored tasks assess the downstream value of the synthesized environments and data for agents facing new evidence, constraints, and required effects.

\end{document}